\documentclass{article}

\usepackage{arxiv}

\usepackage[utf8]{inputenc} 
\usepackage[T1]{fontenc}    
\usepackage{hyperref}       
\usepackage{url}            
\usepackage{booktabs}       
\usepackage{amsfonts}       
\usepackage{nicefrac}       
\usepackage{microtype}      
\usepackage{lipsum}		
\usepackage{graphicx}
\usepackage{doi}

\usepackage{xcolor}
\definecolor{lgray}{gray}{0.8}
\usepackage{colortbl}
\usepackage{array}
\usepackage{booktabs}
\usepackage{multirow}
\usepackage{subcaption}
\usepackage{amsthm}
\usepackage{amsmath}
\usepackage{amssymb}
\usepackage{tikz}
\usetikzlibrary{shapes}
\usetikzlibrary{shapes.multipart}
\usetikzlibrary{shapes.misc}
\usetikzlibrary{fit}
\usetikzlibrary{arrows}
\usetikzlibrary{patterns}
\usetikzlibrary{decorations.markings}
\tikzset{->-/.style={decoration={
			markings,
			mark=at position .65 with {\arrow{triangle 60}}},postaction={decorate}},
	every overlay node/.style={
		draw=white,anchor=north west,
	}
}

\newcommand{\mtt}[1]{\text{\texttt{#1}}}
\newcommand{\kg}{\mathcal{K}}

\newcommand{\lsnn}{\mathcal{L}_{\text{SNN}}}

\title{Discovering alignment relations with Graph Convolutional Networks: a biomedical case study}

\date{}

\author{ \href{https://orcid.org/0000-0002-2017-8426}{\includegraphics[scale=0.06]{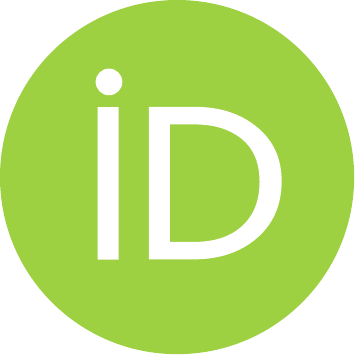}\hspace{1mm}Pierre Monnin} \\
	Universit\'e de Lorraine, CNRS, Inria, LORIA,
	F-54000 Nancy, France \\
	Orange, Belfort, France \\
	\texttt{pierre.monnin@loria.fr} \\
	\And
	\href{https://orcid.org/0000-0002-6100-7519}{\includegraphics[scale=0.06]{orcid.pdf}\hspace{1mm}Chedy Ra\"issi} \\
	Universit\'e de Lorraine, CNRS, Inria, LORIA, 
	F-54000 Nancy, France \\
	Ubisoft, Singapore \\
	\texttt{chedy.raissi@inria.fr} \\
	\And
	Amedeo Napoli \\
	Universit\'e de Lorraine, CNRS, Inria, LORIA, 
	F-54000 Nancy, France \\
	\texttt{amedeo.napoli@loria.fr} \\
	\And
	\href{https://orcid.org/0000-0002-1466-062X}{\includegraphics[scale=0.06]{orcid.pdf}\hspace{1mm}Adrien Coulet} \\
	Universit\'e de Lorraine, CNRS, Inria, LORIA, 
	F-54000 Nancy, France \\
	Inria Paris, F-75012 Paris \\ 
	Centre de Recherche des Cordeliers (UMR1138 Inserm, Universit\'e de Paris, Sorbonne Universit\'e), \\F-75006 Paris, France \\
	\texttt{adrien.coulet@inria.fr} \\
}

\begin{document}
\maketitle

\begin{abstract}
Knowledge graphs are freely aggregated, published, and edited in the Web of data, and thus may overlap. 
Hence, a key task resides in aligning (or matching) their content.
This task encompasses the identification, within an aggregated knowledge graph, of nodes that are equivalent, more specific, or weakly related.
In this article, we propose to match nodes within a knowledge graph by \textit{(i)} learning node embeddings with Graph Convolutional Networks such that similar nodes have low distances in the embedding space, and \textit{(ii)} clustering nodes based on their embeddings, in order to suggest alignment relations between nodes of a same cluster.
We conducted experiments with this approach on the real world application of aligning knowledge in the field of pharmacogenomics, which motivated our study. We particularly investigated the interplay between domain knowledge and GCN models with the two following focuses. 
First, we applied inference rules associated with domain knowledge, independently or combined, before learning node embeddings, and we measured the improvements in matching results.
Second, while our GCN model is agnostic to the exact alignment relations (\textit{e.g.}, equivalence, weak similarity), we observed that distances in the embedding space are coherent with the ``strength'' of these different relations (\textit{e.g.}, smaller distances for equivalences), letting us considering clustering and distances in the embedding space as a means to suggest alignment relations in our case study.
\end{abstract}


\section{Introduction}
\label{section:introduction}

The Semantic Web~\cite{berners2001} offers tools and standards that facilitate the construction of knowledge graphs~\cite{hogan2020} that may aggregate data and elements of knowledge of various provenances. 
The combined use of these scattered elements of knowledge allows access to a larger extent of the available knowledge, which is beneficial to many applications, such as fact-checking or query answering.
For this conjoint use to be possible, one crucial task lies in \textit{matching} units across knowledge graphs or within an aggregated knowledge graph, \textit{i.e.}, finding \textit{alignments} or correspondences between nodes, edges, or subgraphs.
This task is well-studied in the \emph{Ontology Matching} research field~\cite{euzenatS13} and is challenging since knowledge graphs differ in quality, completeness, vocabularies, and languages.
Consequently, different alignment relations may hold between units: some may indicate that two units are equivalent, weakly related, or that one is more specific than the other.

In the present work, we focus on matching specific nodes within an aggregated knowledge graph represented within Semantic Web standards.
We view such a knowledge graph as a directed and labeled multigraph in which nodes represent entities of a world -- also named individuals -- (\textit{e.g.}, places, drugs), literals (\textit{e.g.}, dates, integers), or classes of individuals (\textit{e.g.}, \texttt{Person}, \texttt{Drug}).
It should be noted that we discard litterals from the scope of the present work.
Nodes are linked together through edges defined as triples $\langle \mtt{subject},$ $\mtt{predicate},$ $\mtt{object} \rangle$ in the Resource Description Framework (RDF) format language, where the \texttt{predicate} qualifies the relationship holding between the \texttt{subject} and the \texttt{object} (\textit{e.g.}, \texttt{has\--side\--effect}, \texttt{has-name}).
Entities, classes, and predicates are identified by Uniform Resource Identifiers (URIs).
Knowledge graphs can be associated with ontologies, \textit{i.e.}, formal representations of a domain~\cite{gruber1993translation}, in which classes and predicates are organized in two distinct hierarchies.

\begin{figure}
	\begin{center}
		\includegraphics[scale=0.7]{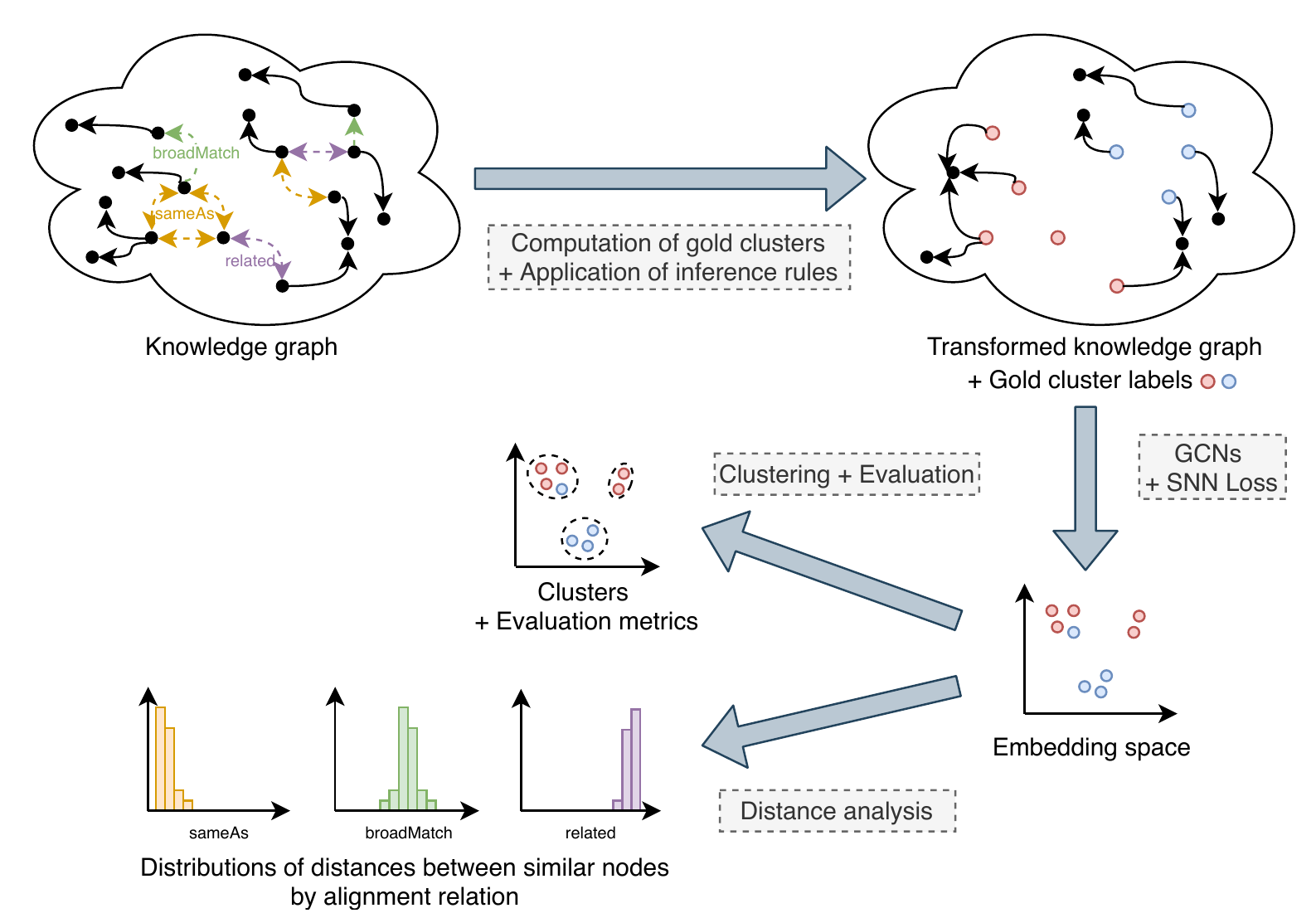}
	\end{center}
	\caption{Outline of our approach. 
		Gold clusters are computed from existing alignments between the nodes to match in the knowledge graph (\textit{e.g.}, \texttt{owl:sameAs}, \texttt{skos:broadMatch}, \texttt{skos:related}, etc.). 
		These alignments are then removed and various inferences rules associated with domain knowledge are applied on the knowledge graph.
		Embeddings of nodes are learned with Graph Convolutional Networks (GCNs) and the Soft Nearest Neighbor (SNN) loss.
		Clustering algorithms are then applied on the embedding space and the resulting clusters are evaluated with regard to the gold clusters.
		A distance analysis is also performed for each alignment relation.}
	\label{fig:approach-outline}
\end{figure}

We propose to match specific individuals that represent n-ary relationships through an approach that combines graph embedding and clustering, outlined in Figure~\ref{fig:approach-outline}. 
Graph embeddings are low-dimensional vectors that represent graph substructures (\textit{e.g.}, nodes, edges, subgraphs) while preserving as much as possible the properties of the graph~\cite{caiZC18}.
More precisely, we learn node embeddings with Graph Convolution Networks (GCNs)~\cite{kipfW17,schlichtkrullKB18} such that similar nodes have a low distance between their embeddings.
We employ graph embeddings since their continuous nature may provide the needed flexibility to cope with the heterogeneous representations of nodes to match~\cite{guha15}.
GCNs compute the embedding of a node by considering the embeddings of its neighbors in the graph.
Hence, nodes with similar neighborhoods will have similar embeddings, what is well-adapted to a structural and relational matching approach~\cite{pangZTT019,wangLLZ18}.

To suggest alignment relations from node embeddings, we apply a clustering algorithm on the embedding space and consider nodes that belong to the same cluster as similar.
The resulting clusters are evaluated by comparison with \textit{gold clusters}, \textit{i.e.}, reference clusters that we aim to reproduce.
We define these gold clusters as groups of nodes linked directly or indirectly by preexisting alignments we obtained from a rule-based method previously published~\cite{monnin2020iccs}.
These pre-existing alignments use five different alignment relations.
For example, nodes may be identical (\texttt{owl:sameAs} links), one may be more specific than the other (\texttt{skos:broadMatch} links), or weakly similar (\texttt{skos:related} links).
Hence, our approach is supervised and requires the preexistence of such alignments. 

Within our approach, we particularly investigated the interplay between GCNs and domain knowledge through the two following aspects.
First, similarly to existing works with different embedding models~\cite{ianaP20}, we applied various inference rules associated with domain knowledge (\textit{e.g.}, class and predicate hierarchies, symmetry and transitivity of predicates), independently or combined, before learning node embeddings, and we measured the improvements or declines in matching results.
Second, we explored how embeddings can differentiate between different types of alignment relations.
We made our GCN model agnostic to these exact relations during learning.
However, we observed that distances between the embeddings of similar nodes are different and coherent with the type and ``strength'' of each alignment relation (\textit{e.g.}, smaller distances for equivalences, larger distances for weak similarities).
Such results allow us to think that distances in the embedding space can be used to suggest alignment relations to connect nodes, in respect with distinct types of similarities.
To the best of our knowledge, our approach is the first one to investigate these aspects in a matching task, combining GCNs and clustering.

\begin{figure}
	\begin{center}
		\begin{tikzpicture}[scale=1,auto=center]
			\node[draw,rounded rectangle,fill=blue!10] (n1) at (-2.2,0.5) {\tiny\texttt{CYP2C9}};
			\node[draw,rounded rectangle,fill=green!10] (n2) at (-2.2,-0.5) {\tiny\texttt{warfarin}};
			\node[draw,rounded rectangle,fill=red!10] (n3) at (3.5,0) {\tiny\texttt{vascular\_disorders}};
			\node[draw,rounded rectangle,fill=yellow!10] (n4) at (0,0) {\tiny\texttt{pgr\_1}};
			
			\draw[-triangle 45] (n1) -- (n4) node[midway,sloped,above] {\tiny\texttt{causes}};
			\draw[-triangle 45] (n2) -- (n4) node[midway,sloped,below] {\tiny\texttt{causes}};
			\draw[-triangle 45] (n4) -- (n3) node[midway,sloped,above] {\tiny\texttt{causes}};
		\end{tikzpicture}
	\end{center}
	\caption{Representation of a PGx relationship between gene \texttt{CYP2C9}, drug \texttt{warfarin} and phenotype \texttt{vascular\_disorders}.
		This relationship is reified through the individual \texttt{pgr\_1}, connecting its components through the \texttt{causes} predicate.}
	\label{fig:pgx-relationship}
\end{figure}
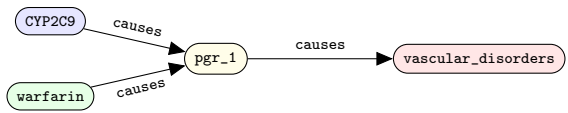

Our approach based on GCNs was motivated by the need to align pharmacogenomic (PGx) knowledge that we previously aggregated in a knowledge graph named PGxLOD~\cite{monninLHRTJNC19}. 
The biomedical domain of PGx studies the influence of genetic factors on drug response phenotypes.
As an example, Figure~\ref{fig:pgx-relationship} depicts the relationship \mtt{pgr\_1}, which states that patients treated with warfarin may experience vascular disorders because of variations in the CYP2C9 gene.
PGx knowledge originates from distinct sources: reference databases such as PharmGKB~\cite{whirlMDHGSTAKT12}, biomedical literature, or the mining of Electronic Health Records of hospitals.
Consequently, there is an interest in matching these sources to obtain a consolidated view of the PGx knowledge.
Such a view would certainly be beneficial to precision medicine, which aims at tailoring drug treatments to patients to reduce adverse effects and maximize drug efficacy~\cite{caudie,couletS16}.
Elements of PGx knowledge consist of $n$-ary relationships between drugs, genomic variations, and phenotypes, whereas only binary relations exist in Semantic Web standards.
Thus, PGx relationships in PGxLOD are reified as individual nodes whose neighbors are the involved drugs, genetic factors, and phenotypes (see Figure~\ref{fig:pgx-relationship})~\cite{noy2006defining}.
In this context, matching PGx relationships reduces to matching the nodes resulting from their reification.
By using GCNs, we hope that nodes representing PGx relationships that involve similar drugs, genetic factors, and phenotypes will have similar embeddings since they have similar neighborhoods.

The remainder of this paper is organized as follows.
In Section~\ref{section:related-works}, we outline some works related to node matching in knowledge graphs and graph embeddings.
We detail the core of our matching approach (node embeddings and clustering) in Section~\ref{section:matching-approach}, and how inference rules associated with domain knowledge are considered in Section~\ref{section:reasoning-mechanisms}.
In Section~\ref{section:experiments}, we conduct experiments with this approach on PGxLOD, a large knowledge graph we built that contains 50,435 PGx relationships~\cite{monninLHRTJNC19}.
Finally, we discuss our results and conclude in Section~\ref{section:discussion} and~\ref{section:conclusion}.

\section{Related work}
\label{section:related-works}

\subsection{Matching}

Numerous papers exist about knowledge graph matching. 
The interested reader could refer to the book of Euzenat and Shvaiko~\cite{euzenatS13} for a formalization of the matching task, and a detailed presentation of the main methods. 
In the following, we focus on graph embedding techniques.
Such techniques have been successfully applied on knowledge graphs for various tasks such as node classification, link prediction, or node clustering~\cite{wangMWG17,caiZC18}. 
Interestingly, the task of matching nodes can be alternatively tackled as a link prediction task (\textit{i.e.}, predicting alignments between nodes) or as a node clustering task (\textit{i.e.}, grouping similar nodes into clusters).
Here, we choose the node clustering approach.

\subsection{Graph embedding}

Existing papers about graph embedding differ in the considered type of graphs (\textit{e.g.}, homogeneous graphs, heterogeneous graphs such as knowledge graphs) or in the embedding techniques (\textit{e.g.}, matrix factorization, deep learning with or without random walk). The survey of Cai et al.~\cite{caiZC18} presents a taxonomies of graph embedding problems and techniques. 
Hereafter, a few specific examples are detailed but a more thorough overview can be found in the following surveys~\cite{caiZC18,nickel0TG16,wangMWG17}.
Some approaches are translational.
For example, TransE~\cite{bordesUGWY13} computes for each triple $\langle s, p, o \rangle$ of a knowledge graph, embeddings $h_{s}$, $h_p$, $h_o$, such that $h_s + h_p \approx h_o$, \textit{i.e.}, the translation vector from the subject to the object of a triple corresponds to the embedding of the predicate. 
This approach is adapted for link prediction but, according to the authors, it is unclear if it can adequately model relations of distinct arities, such as \textit{1-to-Many}, or \textit{Many-to-Many}.
Other approaches use random walks in the knowledge graph.
For example, RDF2Vec~\cite{ristoskiP16} first extracts, for each node, a set of sequences of graph sub-structures starting from this node. 
Elements in these sequences can be edges, nodes, or subtrees. 
Then, sequences feed the word2vec model that compute embeddings for each element in a sequence by either maximizing the probability of an element given the other elements of the sequence (Continuous Bag of World architecture) or maximizing the probability of the other elements given the considered element (Skip-gram architecture).

\subsection{Graph Convolutional Networks (GCNs)}

The approach adopted in this article is based on Graph Convolutional Networks (GCNs).
GCNs have been introduced for semi-supervised classification over graphs~\cite{kipfW17} and extended for entity classification and link prediction in knowledge graphs~\cite{schlichtkrullKB18}. 
In contrast with TransE and RDF2Vec that work at the triple and sequence levels, GCNs compute the embedding of a node by considering its neighborhood in the graph. 
Hence, as aforementioned, we believe GCNs are well-suited for our application of matching reified $n$-ary relationships since similar relationships have similar neighborhoods.
Other existing works rely on this assumption that similar nodes have similar neighborhoods and use GCNs for their matching.
For example, Wang et al.~\cite{wangLLZ18} propose to align cross-lingual knowledge graphs by using GCNs to learn node embeddings such that nodes representing the same entity in different languages have close embeddings.
Pang et al.~\cite{pangZTT019} use the same approach to align two knowledge graphs, but introduce an iterative aspect.
Some newly-aligned entities are selected and used when learning embeddings in the next iteration.
To avoid introducing false positive alignments, the newly-aligned entities are selected with a distance-based criteria proposed by the authors.
Interestingly, the two previous approaches take into account literals in the embedding process and use the \textit{triplet loss}, also used by TransE.
On the contrary, in our work, we discard literals and use the \textit{Soft Nearest Neighbor loss}~\cite{frosstPH19} to consider all positive and negative examples instead of sampling\footnote{GCNs and the Soft Nearest Neighbor loss are further detailed in Subsection~\ref{subsection:learning-embeddings}}.

\subsection{Graph embedding and domain knowledge}

However, previous methods do not consider inference rules associated with domain knowledge represented in knowledge graphs on the contrary of recent papers~\cite{paulheim18}.
For example, Iana and Paulheim~\cite{ianaP20} evaluate the RDF2Vec embedding model when inferred triples associated with  subproperties, symmetry, and transitivity of predicates are added to the knowledge graph.
Interestingly, the addition of inferred triples seems to degrade the performance of RDV2Vec embeddings in downstream applications (\textit{e.g.}, regression, classification).
Instead of materializing inferred triples into the knowledge graph, d'Amato et al.~\cite{dAmatoQF21} propose to inject domain knowledge in the learning process by defining specific loss functions and scoring functions for triples.
Logic Tensor Networks~\cite{serafiniG16} learn groundings of logical terms and logical clauses.
The grounding of a logical term consists in a vector of real numbers (\textit{i.e.}, an embedding) and the grounding of a logical clause is a real number in the interval $\left[0, 1\right]$ (\textit{i.e.}, the confidence in the truth of the clause).
The learning process aims at minimizing the satisfiability error of a set of clauses, while ensuring the logical reasoning. 
This work can interestingly be compared to graph embeddings if knowledge graphs are considered in their logical form, \textit{i.e.}, considering nodes as logical terms and edges linking two nodes as logical formulae. 
Alternatively, Wang et al.~\cite{wangHLP19} propose an hybrid attention mechanism named ``Logic Attention Network'' (LAN) in embedding approaches for link prediction.
LAN combines a mechanism based on logical rules and a neural network mechanism.
The rule-based mechanism weights neighbors by promoting those linked by a predicate that has been found to strongly imply the predicate of the link to predict.
Besides implications between predicates, more complex logical rules can be associated with knowledge graphs through ontologies.
That is why Guti{\'{e}}rrez{-}Basulto and Schockaert~\cite{gutierrez-Basulto18} investigate how to ensure logical consistency through geometrical constraints on embedding spaces and if classical embedding techniques respect such constraints.
Similarly, \textsf{OWL2Vec*}~\cite{chenHJHAH21} focus on embedding complex logical constructors as well as the graph structure and literals.

These related works and our preliminary results~\cite{monninRNC19} inspired the present work where we investigate how \textit{(i)} inference rules associated with domain knowledge can improve the performances in node matching and \textit{(ii)} the distance in the embedding space is representative of the type and ``strength'' of alignment relations, and thus can be used to suggest the specific relation to use between matched nodes.

\section{Matching nodes with Graph Convolutional Networks and clustering}
\label{section:matching-approach}

\subsection{Approach outline}

Our approach is outlined in Figure~\ref{fig:approach-outline}.

It takes as input an aggregated knowledge graph $\kg$ and a set $S$ of nodes to match, where $S$ is a subset of the nodes of $\kg$.
This initial selection of the nodes to match is motivated by our biomedical application as we only intend to match nodes that represent reified PGx relationships.
We discard literals and edges incident to literals from $\kg$ and $S$.
Hence, a node is either an entity or a class.
We consider that we have at our disposal \textit{gold clusters}, \textit{i.e.}, sets of nodes from $S$ that are already labeled as similar. 
These gold clusters can have uneven sizes.
We propose to match nodes in $S$ as follows:
\begin{enumerate}
	\item Learn embeddings for all nodes in $\kg$ such that nodes in $S$ labeled as similar (\textit{i.e.}, belonging to the same gold cluster) have smaller distances between their embeddings (Subsection~\ref{subsection:learning-embeddings}).
	\item Apply a clustering algorithm only on the embeddings of nodes from $S$ and consider nodes belonging to the same cluster as similar (Subsection~\ref{subsection:clustering}).
\end{enumerate}

It should be noted that gold clusters can result from another automatic matching method or a manual alignment by an expert.
For example, in Section~\ref{section:experiments}, our gold clusters are computed from alignments semi-automatically obtained with rules manually written by experts~\cite{monnin2020iccs}.
These alignments can use different alignment relations (\textit{e.g.}, equivalence, weak similarity).
We further detail in Subsection~\ref{subsection:gold-clusters} how distinct relations are taken into account in our experiments.

\subsection{Learning node embeddings with Graph Convolutional Networks and the Soft Nearest Neighbor loss}
\label{subsection:learning-embeddings}

To learn embeddings for all nodes in $\kg$, we propose to use Graph Convolutional Networks (GCNs) and the Soft Nearest Neighbor loss.
In the following, we adopt the model, notations, and definitions of Schlichtkrull et al.~\cite{schlichtkrullKB18}. 
As such, $\mathcal{R}$ denotes the set of predicates in the considered knowledge graph $\kg$. 
Given a node $i$ and a predicate $r \in \mathcal{R}$, we denote by $\mathcal{N}_i^r$ the set of nodes reachable from $i$ by an edge labeled by $r$. 

Graph Convolutional Networks (GCNs) can be seen as a message-passing framework of multiple layers, in which the embedding $h_i^{(l+1)}$ of a node $i$ at layer $(l+1)$ depends on the embeddings of its neighbors at level $(l)$, as stated in Eq.~\eqref{eq:gcn}. 
\begin{equation}
	h_i^{(l+1)} = \sigma \Bigg( \sum_{r \in \mathcal{R}} \sum_{j \in \mathcal{N}_i^r} \frac{1}{c_{i,r}} W_r^{(l)} h_j^{(l)} + W_0^{(l)}h_i^{(l)} \Bigg)
	\label{eq:gcn}
\end{equation}
This convolution over the neighboring nodes $j$ of $i$ is computed with a specific weight matrix $W_r^{(l)}$ for each predicate $r \in \mathcal{R}$ and each layer $(l)$. 
The convolution is regularized by a constant $c_{i,r}$, that can be set for each node and each predicate. 
Similarly to Schlichtkrull et al.~\cite{schlichtkrullKB18}, we use $c_{i,r} = |\mathcal{N}_i^r|$.
The weight matrix $W_0^{(l)}$ enables a self-connection, \textit{i.e.}, the embedding of $i$ at layer $(l+1)$ also depends on its embedding at layer $(l)$.
$\sigma$ is a non-linear function such as $\mathrm{ReLU}$ or $\mathrm{tanh}$.

The number of predicates in $\kg$ can lead to a high number of parameters $W^{(l)}_r$ to optimize.
To ensure the scalability of our approach and reduce the number of parameters to optimize, we use the basis-decomposition proposed by Schlichtkrull et al.~\cite{schlichtkrullKB18}.
Hence, each $W^{(l)}_r$ is decomposed as follows:
\begin{equation}
	W^{(l)}_r = \sum_{b = 1}^{B} a^{(l)}_{rb} V^{(l)}_b
	\label{eq:basis-decomposition}
\end{equation}
For each level $(l)$, $B$ matrices $V_b^{(l)} \in \mathbb{R}^{d^{(l+1)}\times d^{(l)}}$ and $|\mathcal{R}|\times B$ coefficients $a^{(l)}_{rb} \in \mathbb{R}$ are learned, where $d^{(l)}$ and $d^{(l+1)}$ denote the dimension of embeddings at level $(l)$ and level $(l+1)$ respectively.
Then, each $W^{(l)}_r$ is computed as a linear combination of matrices $V^{(l)}_b$ and coefficients $a^{(l)}_{rb}$.
As only these coefficients depend on predicates $r$, the number of parameters to learn is reduced.

Recall that our objective is to cluster similar nodes, which differs from previous applications of GCNs (\textit{e.g.}, node classification, link prediction~\cite{kipfW17,schlichtkrullKB18,wangLLZ18}).
Hence, we propose to train GCNs from scratch by minimizing the Soft Nearest Neighbor (SNN) loss which, to the best of our knowledge, has never been used with GCN models before.
This loss was defined by Frosst et al.~\cite{frosstPH19} and is presented in Eq.~\eqref{eq:snn-loss}, 
\begin{equation}
	\lsnn = -\cfrac{1}{|N|} \sum_{i \in N} \log \left(\cfrac{\sum\limits_{\substack{j \in N\\j\neq i \\ Y_i = Y_j}} e^{-\frac{||h_i - h_j||^2}{T}}}{\sum\limits_{\substack{k \in N\\k\neq i}} e^{-\frac{||h_i - h_k||^2}{T}}}\right)
	\label{eq:snn-loss}
\end{equation}
The input of the SNN loss consists of:
\begin{itemize}
	\item A set $N$ of nodes belonging to the gold clusters (see Subsection~\ref{subsection:node-embeddings}).
	\item A set $Y$ of labels for nodes in $N$. These labels corresponds to the assignments of nodes in $N$ to the gold clusters.
	\item A temperature $T$.
	\item Embeddings $h$ of nodes. These embeddings are the output of the last layer of the GCN model.
\end{itemize}
Minimizing the SNN loss corresponds to minimizing intra-cluster distances and maximizing inter-cluster distances for the gold clusters of nodes in $N$.
The temperature $T$ determines how distances influence the loss.
Indeed, distances between widely separated embeddings are taken into account when $T$ is large whereas only distances between close embeddings are taken into account when $T$ is small.
To avoid $T$ as an hyperparameter of the model, we adopt the same learning procedure as Frosst et al.~\cite{frosstPH19}: $T$ is initialized to a predefined value and is optimized by learning $\frac{1}{T}$ as a model parameter.

The computation of $\lsnn$ (Eq.~\eqref{eq:snn-loss}) considers all positive and negative examples from $N$.
Indeed, distances between nodes with the same label are minimized (\textit{i.e.}, positive examples) whereas distances between nodes with different labels are maximized (\textit{i.e.}, negative examples).
However, it is noteworthy that $\kg$ is based on the \textit{Open World Assumption}.
Hence, nodes with different labels are regarded as dissimilar (\textit{i.e.}, negative examples) while their (dis)similarity may only be unknown.

This step enables to learn embeddings for all nodes in $\kg$ such that distances between embeddings of identified similar nodes (in $N$ in $\lsnn$) are low.
Note that, in this learning procedure, the semantics of relations in $\kg$ is not taken into account. 
We propose to take into account this semantics with a preprocessing step presented in Section~\ref{section:reasoning-mechanisms}.

\subsection{Matching nodes by clustering their embeddings}
\label{subsection:clustering}

After embeddings of all nodes in the graph have been output by the last layer of the GCN, we perform a clustering on embeddings $h_i$ for all nodes $i \in S$, \textit{i.e.}, all nodes to check for matching.
Nodes assigned to one same cluster are considered as similar and connected with an alignment relation. Clusters are evaluated with regard to gold clusters.

We conduct comparative experiments with three distinct clustering algorithms presented in Table~\ref{tab:clustering-algorithms}, in regards with three classical metrics presented in Table~\ref{tab:performance-metrics}.
Within the large set of existing algorithms, our choice has been guided by the constraints of our task that requires to handle an important number of clusters, potentially large, and with uneven sizes (see Subsection~\ref{subsection:gold-clusters} and Figure~\ref{fig:gold-clusters} for the sizes of gold clusters computed on PGxLOD). We decided to arbitrarily limit ourselves to three algorithms, but decided to opt for algorithms that cover some diversity in the various family of algorithms.
Our three algorithms differ in their parameters: in particular they require either the number of clusters to find (Ward and Single) or the minimum size of clusters (OPTICS).
We used our set of gold clusters to set these parameters.
Considering these distinct algorithms allows us to evaluate the influence of inference rules in different settings (see Section~\ref{section:reasoning-mechanisms}).

\begin{table*}
	\caption{Clustering algorithms applied on the embeddings of nodes in $S$. Nodes that belong to the same predicted cluster are considered as similar.}
	\label{tab:clustering-algorithms}
	\begin{center}
		\begin{tabular}{llm{10.4cm}}
			\hline
			Algorithm & Parameter & Description \\
			\hline
			Ward & Number of clusters to find & Hierarchical clustering algorithm that successively merges clusters by minimizing the variance of merged clusters  \\
			Single & Number of clusters to find & Hierarchical clustering algorithm that successively merges clusters whose distance between their closest observations is minimal \\
			OPTICS~\cite{ankerstBKS99} & Minimum size of clusters & Algorithm that finds zones of high density and expand clusters from them \\
			\hline
		\end{tabular}
	\end{center}
\end{table*}

\begin{table*}
	\caption{Performance metrics used to compare the clusters predicted by  the algorithms presented in Table~\ref{tab:clustering-algorithms} with gold clusters.}
	\label{tab:performance-metrics}
	\begin{center}
		\begin{tabular}{m{2.4cm}ccm{10.4cm}}
			\hline
			Metric & Abbr. & Domain & Description \\
			\hline
			Unsupervised Clustering Accuracy & ACC & $\left[0,1\right]$ & Counts nodes whose predicted cluster label is the same as their gold cluster label divided by the total number of nodes.
			As labels may be permuted between predicted and gold clusters, the mapping with the best ACC is used. \\
			Adjusted Rand Index & ARI & $\left[-1,1\right]$ & Considers all pairs of nodes and counts those whose nodes are assigned to the same or different clusters both in predicted and gold clusters.
			ARI is equal to $0$ for a random labeling, and equal to $1$ for a perfect labeling (up to a permutation).
			ARI is adjusted for chance. \\ 
			Normalized Mutual Information & NMI & $\left[0,1\right]$ & Measures the mutual information between the predicted and gold clusters, normalized by the entropy of both types of clusters.
			NMI is equal to $1$ for a perfect labeling (up to a permutation). \\
			\hline
		\end{tabular}
	\end{center}
\end{table*}

\section{Evaluating the influence of applying inference rules associated with domain knowledge}
\label{section:reasoning-mechanisms}

Semantic Web knowledge graphs are represented within formalims such as Description Logics~\cite{baader2003} that are equipped with inference rules.
Hence, we propose to evaluate the improvements in the results of our matching approach (detailed in Section~\ref{section:matching-approach}) when considering such inference rules, independently or combined.
Here, we only consider the inference rules associated with the following logic axioms: class and predicate assertions, equivalence axioms between entities or classes, subsumption axioms between classes or predicates, and axioms defining predicate inverses.
We consider this limited set of inference rules because they are the only ones actionable in PGxLOD, the knowledge graph that motivated our study.
Accordingly, we generate six different graphs ($\mathcal{G}_{0-5}$) by running over $\kg$ these inference rules until saturation.
This inference and saturation process is implemented in a Python script, without the use of an inference engine in part because at this stage the graph is in the format of the GCN library and in part because this facilitates the independent activation of inference rules that is required by our experiment.
Soundness and completeness of this inference and saturation process were carefully checked but scalability was not tested on knowledge graphs larger than PGxLOD.
We then test our approach on these six graphs that are summarized in Table~\ref{tab:reasoning-mechanisms} and further described below.

$\mathcal{G}_0$ constitutes the baseline in which no inference rules are run and with the systematic addition of abstract inverses.
Indeed, Schlichtkrull et al.~\cite{schlichtkrullKB18} consider that for every predicate $r \in \mathcal{R}$, there exists an inverse $r_{\text{inv}} \in \mathcal{R}$.
Thus, for every $r \in \mathcal{R}$, we add an abstract inverse $r_{\text{inv}} \in \mathcal{R}$ such that its adjacency matrix represents the inverse of $r$.
This addition of abstract inverses is performed in all other graphs, except when explicitly stated otherwise.
$\mathcal{G}_1$ results from the contraction of \texttt{owl:sameAs} edges.
Indeed, in $\kg$, several nodes representing the same entity can co-exist. 
In this case, they may be linked (directly or indirectly) by \texttt{owl:sameAs} edges and should be considered as one, which is enabled by this contraction.
It is noteworthy that the standard transformation actually consists in propagating edges to equivalent nodes instead of merging them.
However, in our case study, 
the merging transformation was prefered because it is aligned with the architecture of GCNs that corresponds to the topological structure of the graph.
In $\mathcal{G}_2$, we do not always add abstract inverses but consider definitions of inverses and symmetry of predicates instead. 
That is to say:
\begin{itemize}
	\item[(i)] For a predicate $r_1$ defined as symmetric (\textit{i.e.}, $r_1 \equiv r_1^{-1}$), we do not add an abstract inverse $r_{1\ \text{inv}}$ and complete its adjacency matrix to ensure its symmetry.
	\item[(ii)] For a predicate $r_2$ that has a defined inverse $r_3$ (\textit{i.e.}, $r_3 \equiv r_2^{-1}$), we do not add an abstract inverse $r_{2\ \text{inv}}$ and complete their adjacency matrices to ensure they represent inverse predicates.
	\item[(iii)] Otherwise, for a predicate $r_4$ that neither is symmetric nor have a defined inverse, we add an abstract inverse $r_\text{4\ inv}$ such that its adjacency matrix represents the inverse of $r_4$.
\end{itemize} 
$\mathcal{G}_3$ takes into account the hierarchy of predicates. 
Indeed, if a predicate $r_1$ is a subpredicate of $r_2$ (\textit{i.e.}, $r_1 \sqsubseteq r_2$) and a triple $\langle i, r_1, j\rangle$ exists, then we make sure the triple $\langle i, r_2, j\rangle$ also exists in the graph.
This completion is performed by considering the transitive closure of the subsumption relation $\sqsubseteq$.
That is to say, if $r_1 \sqsubseteq r_2$ and $r_2 \sqsubseteq r_3$, we also consider $r_1 \sqsubseteq r_3$.
Similarly, $\mathcal{G}_4$ completes \texttt{type} edges based on the hierarchy of ontology classes defined by \texttt{subClassOf} edges.
Hence, if $\langle i, \mtt{type}, j \rangle$ and $\langle j, \mtt{subClassOf}, k\rangle$ exist in the graph, then we ensure that $\langle i, \mtt{type}, k \rangle$ is also in the graph.
Here again, \texttt{subClassOf} edges are considered by computing their transitive closure.
Finally, $\mathcal{G}_5$ is the graph resulting from all transformations from $\mathcal{G}_1$ to $\mathcal{G}_4$.

\begin{table*}
	\caption{Visual summary of the transformations of $\kg$ to evaluate the influence of the application of inference rules associated with domain knowledge on node matching. $\mathcal{G}_0$ is the baseline that corresponds to no inference rules being run and the systematic addition of abstract inverses.}
	\label{tab:reasoning-mechanisms}
	\begin{center}
		\begin{tabular}{>{\centering\arraybackslash} m{1.1cm} >{\centering\arraybackslash} m{6.5cm}  >{\centering\arraybackslash}m{6.5cm}}
			\hline
			Graph & Before & After \\
			\hline
			$\mathcal{G}_0$ & \includegraphics[scale=0.8]{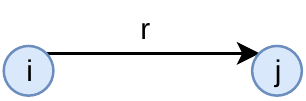} & \includegraphics[scale=0.8]{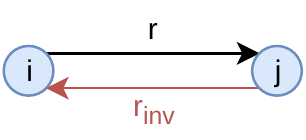} \\
			\hline
			$\mathcal{G}_1$ & \includegraphics[scale=0.8]{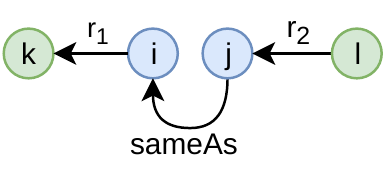} & \includegraphics[scale=0.8]{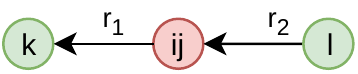}\\
			\hline
			& \includegraphics[scale=0.8]{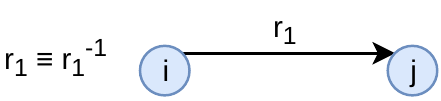} & \includegraphics[scale=0.8]{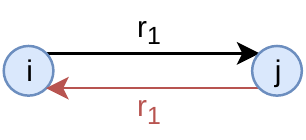} \\
			$\mathcal{G}_2$& \includegraphics[scale=0.8]{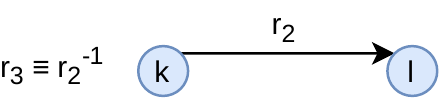} & \includegraphics[scale=0.8]{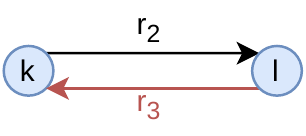} \\
			& \includegraphics[scale=0.8]{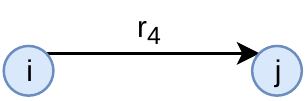} & \includegraphics[scale=0.8]{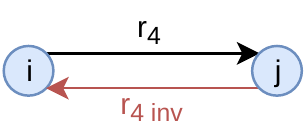} \\
			\hline
			$\mathcal{G}_3$ & \includegraphics[scale=0.8]{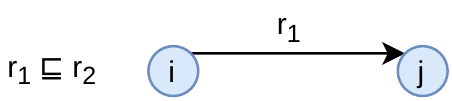} & \includegraphics[scale=0.8]{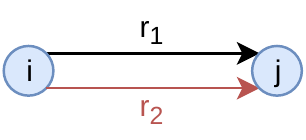}\\
			\hline
			$\mathcal{G}_4$ & \includegraphics[scale=0.8]{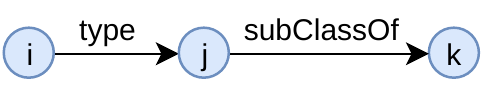} & \includegraphics[scale=0.8]{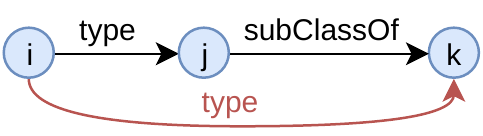} \\
			\hline
			$\mathcal{G}_5$ & \multicolumn{2}{c}{All transformations from $\mathcal{G}_1$ to $\mathcal{G}_4$} \\
			\hline
		\end{tabular}
	\end{center}
\end{table*}

\section{Experiments}
\label{section:experiments}

We conducted experiments with PGxLOD\footnote{\url{https://pgxlod.loria.fr}}, a large knowledge graph about pharmacogenomics (PGx) that we previously built and that motivated this study~\cite{monninLHRTJNC19}.
Our approach is implemented in Python, using PyTorch and the Deep Graph Library for learning embeddings, and scikit-learn for clustering.
Our code is available on GitHub\footnote{\url{https://github.com/pmonnin/gcn-matching}}.

\subsection{Knowledge graph and gold clusters of similar nodes}
\label{subsection:gold-clusters}

PGxLOD presents several characteristics needed in the scope of our study.
First, PGxLOD contains nodes whose matching is well-adapted to a structure-based approach such as ours.
Additionally, alignments are expected to be found between these nodes.
Indeed, PGxLOD contains 50,435 PGx relationships resulting from:
\begin{itemize}
	\item an automatic extraction from the reference data\-base PharmGKB;
	\item an automatic extraction from the biomedical literature;
	\item a manual representation of 10 studies made from Electronic Health Records of hospitals.
\end{itemize}
Alignments are expected to be found between such relationships since, for example, PharmGKB is manually curated by experts after a literature review.
Recall that PGx relationships are $n$-ary, and thus they are reified as nodes, as illustrated in Figure~\ref{fig:pgx-relationship}~\cite{noy2006defining}.
Hence, nodes representing these relationships form our set $S$ of nodes to match.
The reification process entails that neighbors of such nodes are the drugs, genetic factors, and phenotypes involved in the relationships.
Consequently, similar relationships have similar neighborhoods, which makes a structure-based approach such as ours well-adapted for their matching.

Second, PGxLOD contains \texttt{owl:sameAs} edges (or equivalence axioms), which makes possible  the transformation represented in $\mathcal{G}_1$.
Indeed, PGxLOD integrates several Linked Open Data sets: ClinVar, DrugBank, SIDER, DisGeNET, PharmGKB, and CTD.
These LOD sets contain facts describing components of PGx relationships (\textit{i.e.}, drugs, phenotypes, and genetic factors).
Several LOD sets may describe the same entities and we know it explicitly, \textit{i.e.}, some nodes belonging to different LOD sets are linked with \texttt{owl:sameAs} edges.
For example, this could be the case of a drug represented both in PharmGKB and DrugBank.
Thus, we can apply the \texttt{owl:sameAs} identification.

Third, PGxLOD contains subsumption axioms between classes and between predicates, which makes possible the transformations represented in $\mathcal{G}_3$ and $\mathcal{G}_4$.
Indeed, PGxLOD includes the ATC, MeSH, PGxO, and ChEBI ontologies.

\begin{figure*}
	\begin{center}
		\begin{subfigure}{0.32\textwidth}
			\begin{center}
				\includegraphics[width=\textwidth]{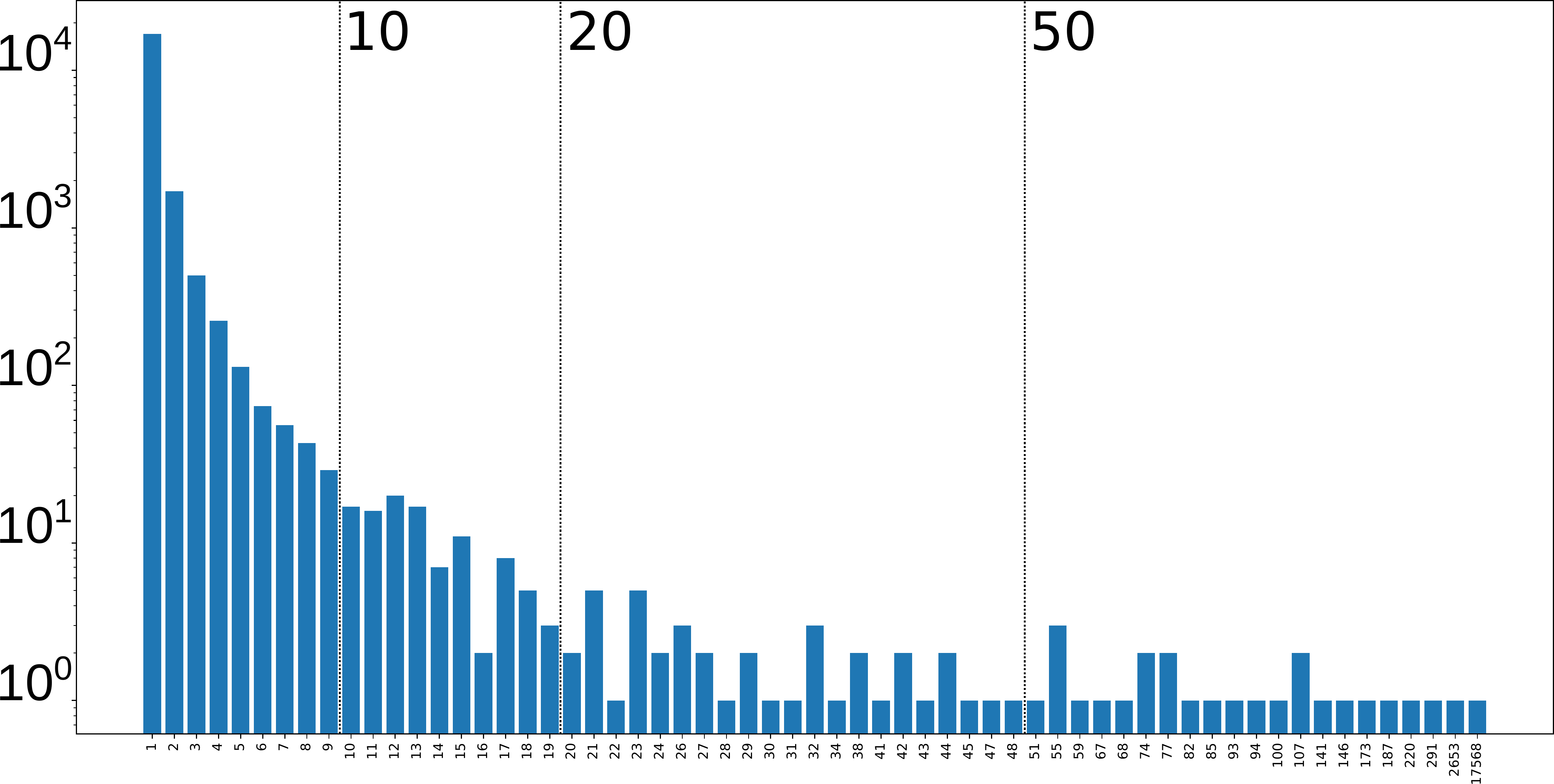}
			\end{center}
			\caption{$\mathcal{C}_0$ ($\max=17,568$)}
		\end{subfigure}
		\begin{subfigure}{0.32\textwidth}
			\begin{center}
				\includegraphics[width=\textwidth]{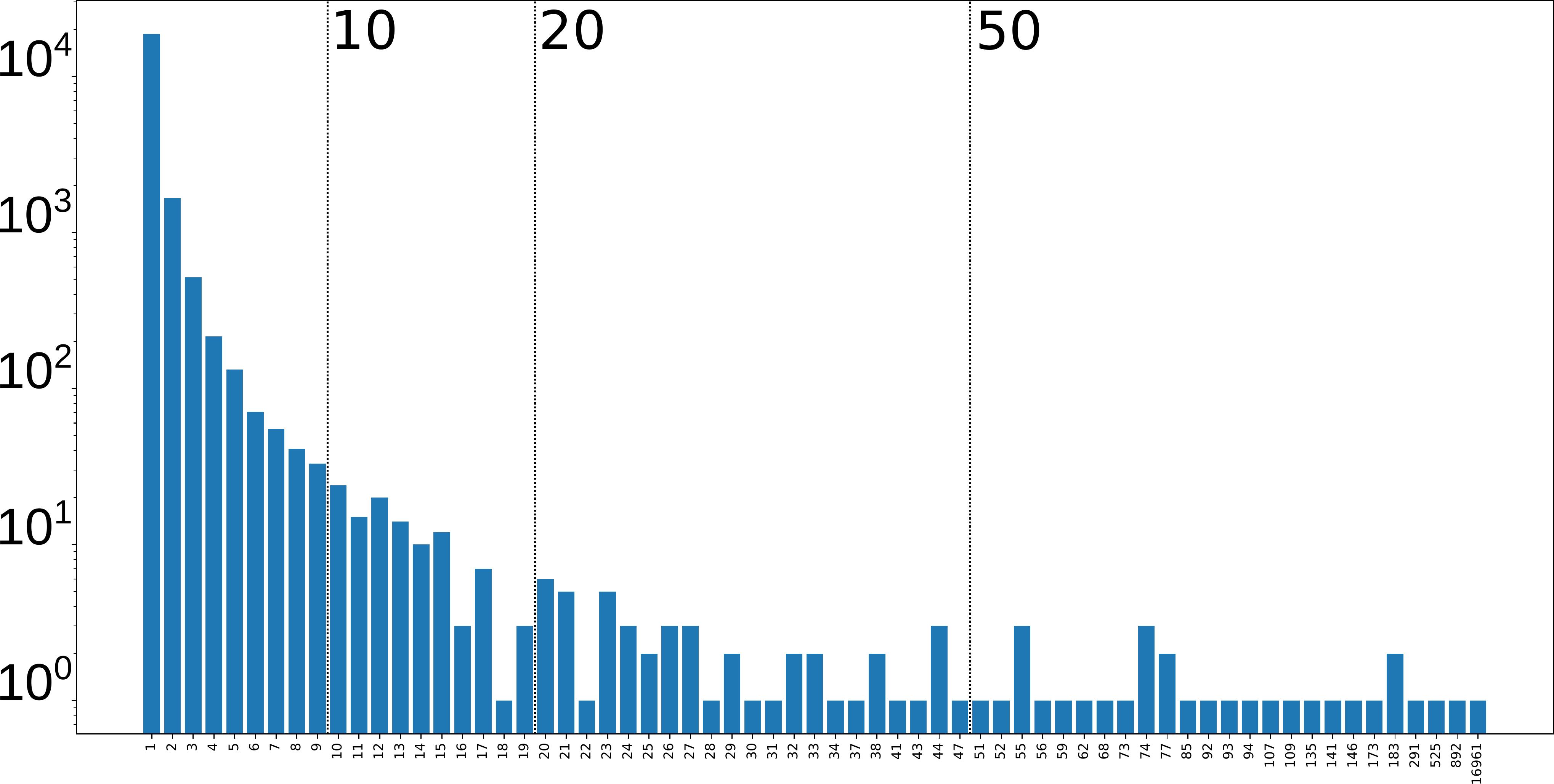}
			\end{center}
			\caption{$\mathcal{C}_1$ ($\max=16,961$)}
		\end{subfigure}
		\begin{subfigure}{0.32\textwidth}
			\begin{center}
				\includegraphics[width=\textwidth]{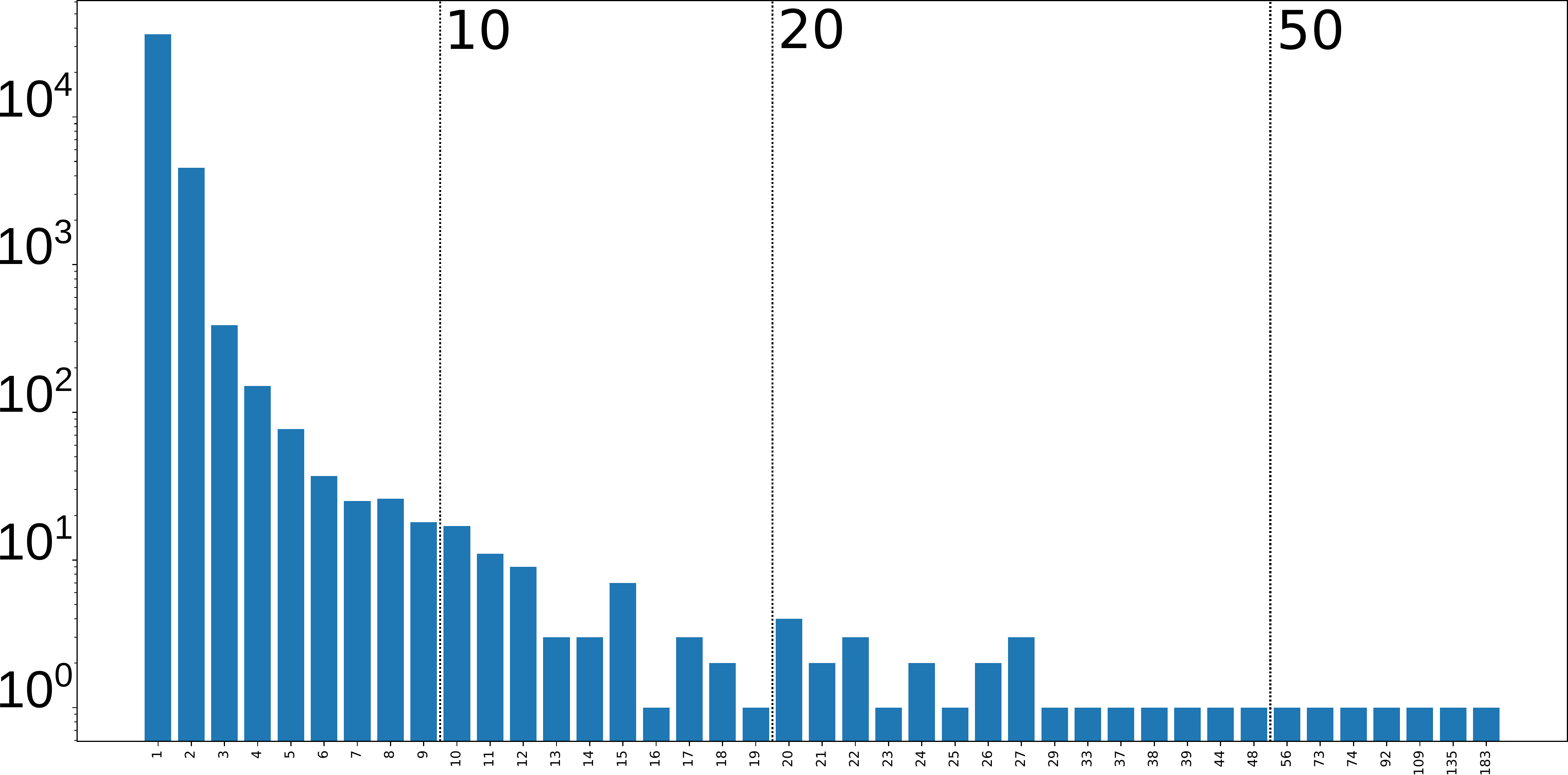}
			\end{center}
			\caption{$\mathcal{C}_2$ ($\max=183$)}
		\end{subfigure}
		
		\vspace{2em}
		
		\begin{subfigure}{0.32\textwidth}
			\begin{center}
				\includegraphics[width=\textwidth]{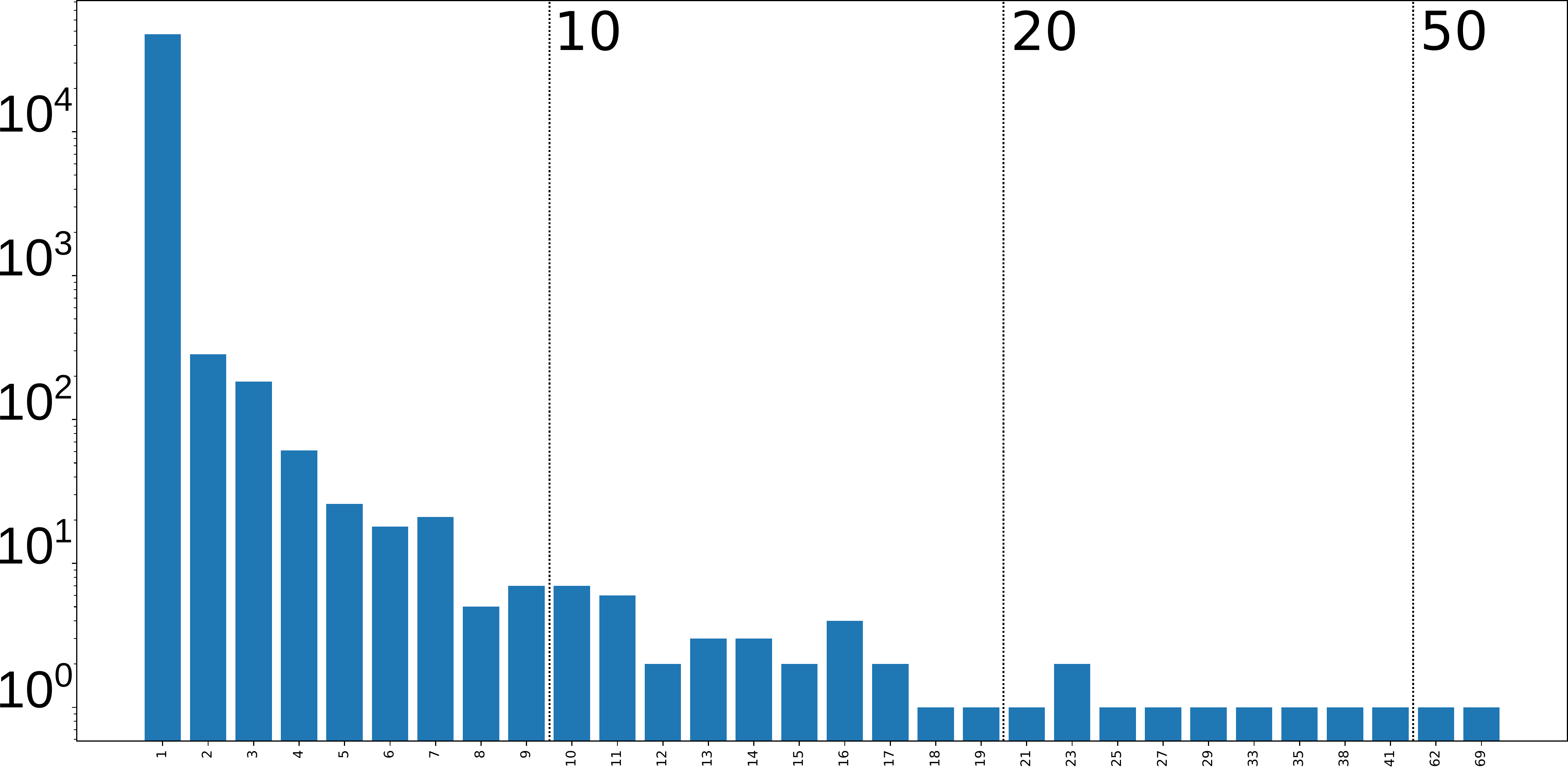}
			\end{center}
			\caption{$\mathcal{C}_3$ ($\max=69$)}
		\end{subfigure}
		\begin{subfigure}{0.32\textwidth}
			\begin{center}
				\includegraphics[width=\textwidth]{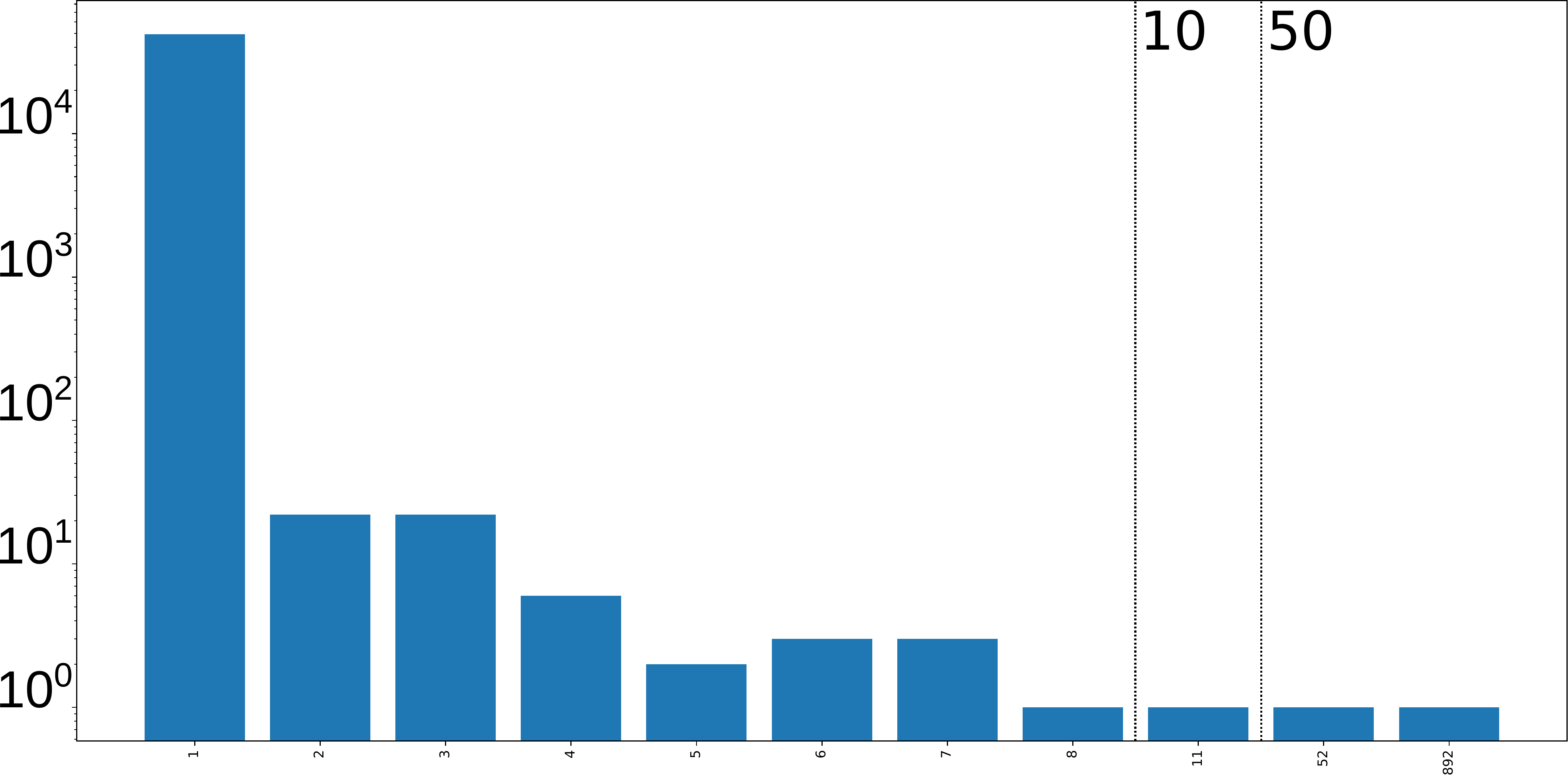}
			\end{center}
			\caption{$\mathcal{C}_4$ ($\max=892$)}
		\end{subfigure}
		\begin{subfigure}{0.32\textwidth}
			\begin{center}
				\includegraphics[width=\textwidth]{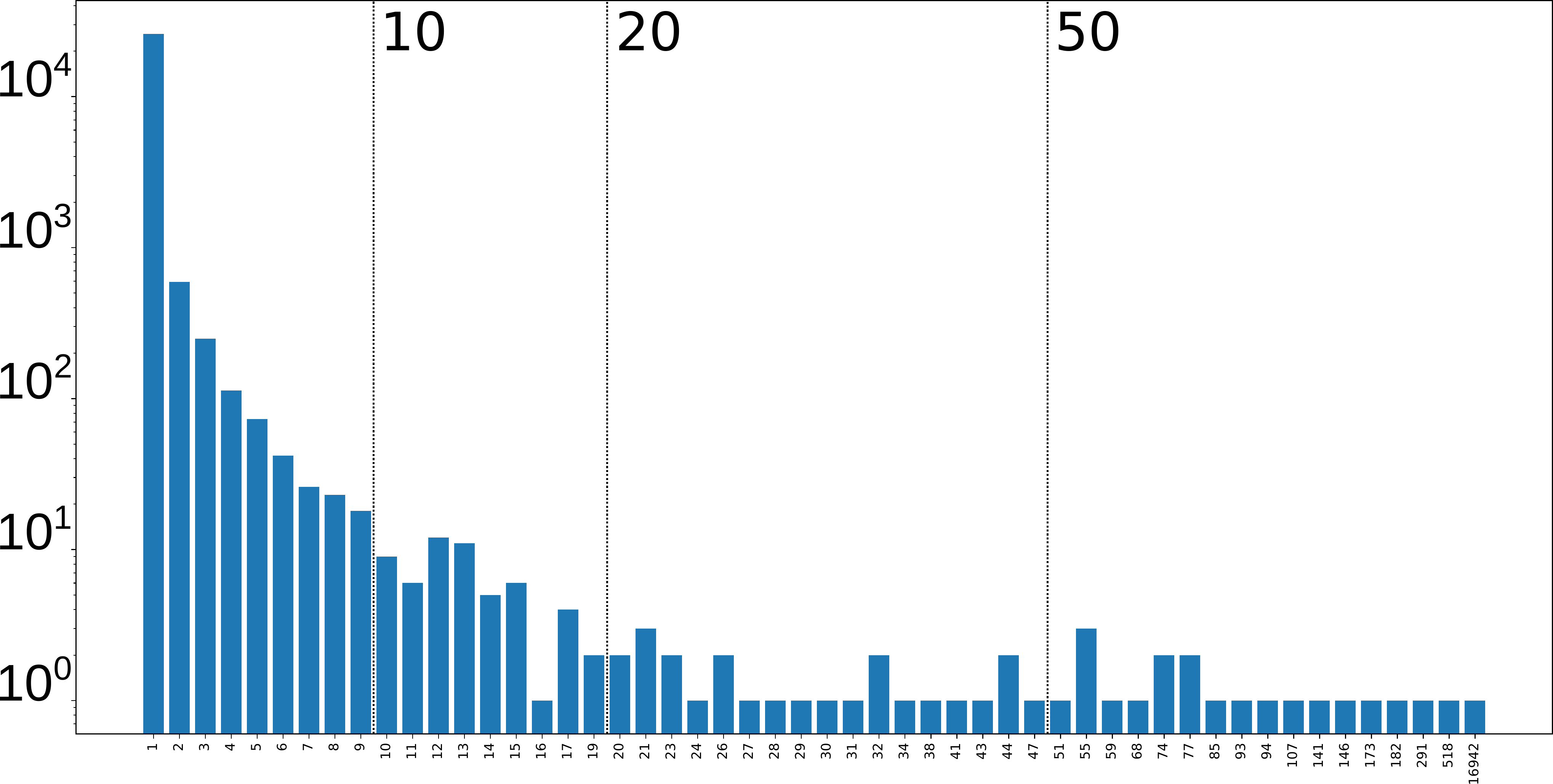}
			\end{center}
			\caption{$\mathcal{C}_5$ ($\max=16,942$)}
		\end{subfigure}
		
		\vspace{2em}
		
		\begin{subfigure}{0.32\textwidth}
			\begin{center}
				\includegraphics[width=\textwidth]{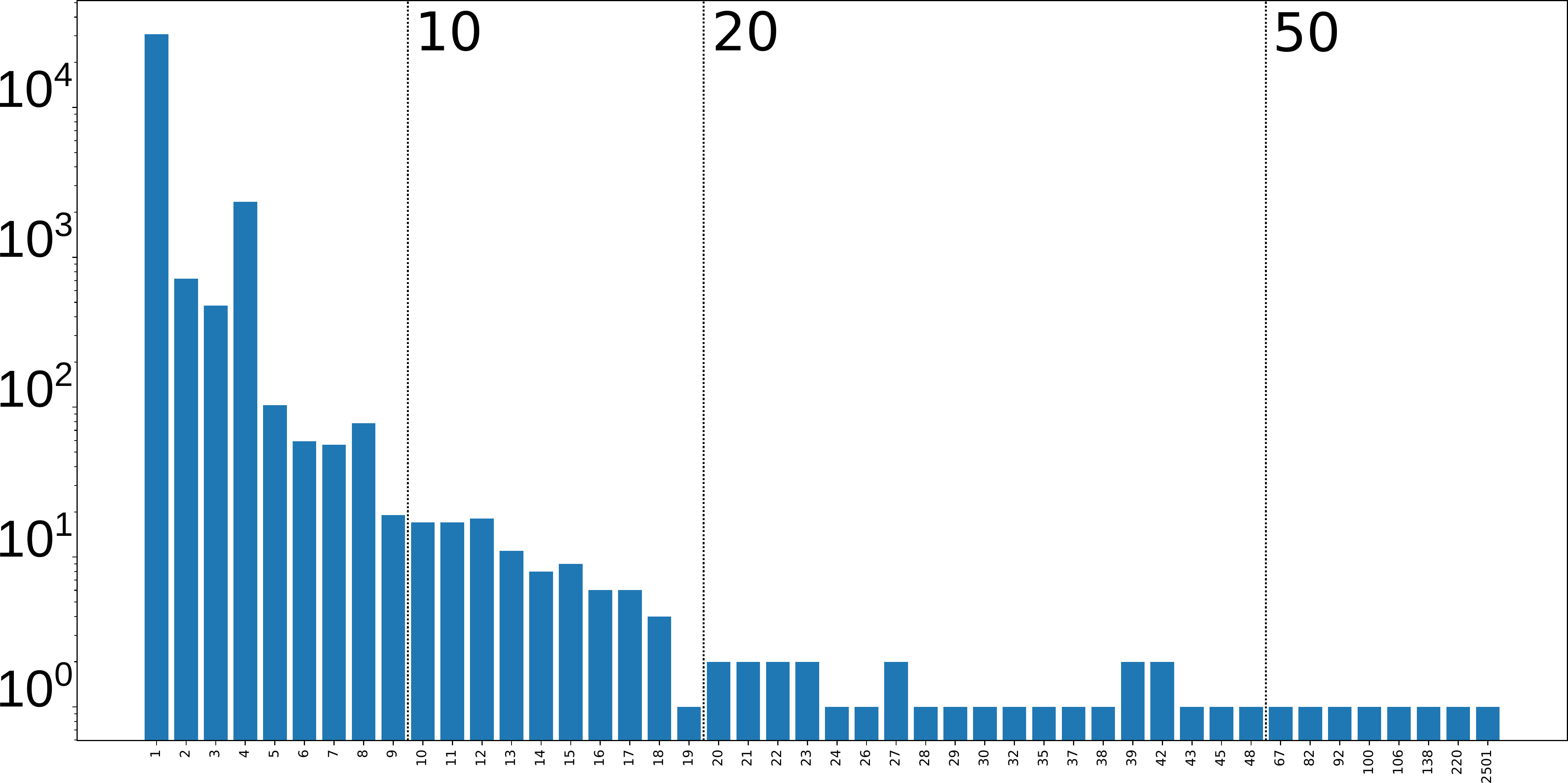}
			\end{center}
			\caption{$\mathcal{C}_6$ ($\max=2,501$)}
		\end{subfigure}
	\end{center}
	\caption{Number of gold clusters (y-axis) by size (x-axis) for each gold clustering. 
		The $\max$ value is the maximum size of gold clusters (in terms of number of nodes).
		The minimum size is 1 for every gold clustering.
		Only gold clusters larger than 10, 20, and 50 nodes are later used to compute performance metrics.
		Gold clusterings are defined in Table~\ref{tab:clustering-links}.}
	\label{fig:gold-clusters}
\end{figure*}

Fourth, some PGx relationships in $S$ are already labeled as similar through alignments resulting from the application of five matching rules previously published~\cite{monnin2020iccs}.
These alignments use the five following alignment relations: \texttt{owl:\-same\-As}, \texttt{skos:\-close\-Match}, \texttt{skos:\-re\-la\-ted\-Match}, \texttt{skos:\-re\-la\-ted}, and \texttt{skos:\-broad\-Match}.
Alignments using \texttt{owl:sameAs} and \texttt{skos:\-close\-Match} indicate strong similarities, whereas \texttt{skos:\-re\-la\-ted\-Match} and \texttt{skos:re\-la\-ted} indicate weaker similarities.
Alignments using \texttt{skos:\-broad\-Match} indicate that a PGx relationship is more specific than another.
These alignments are removed before running inference rules over \(\kg\), learning embeddings, and clustering.
However, they allow to compute \emph{gold clusters}, \textit{i.e.}, sets of nodes that are considered as similar since they are directly or indirectly connected through alignments.
We propose the different \textit{gold clusterings} detailed in Table~\ref{tab:clustering-links} (named \(\mathcal{C}_0\) to \(\mathcal{C}_6\)).
They variously consider the five alignment relations when computing gold clusters to evaluate our approach in different settings (\textit{e.g.}, all the different alignment relations in \(\mathcal{C}_0\), only symmetric relations in \(\mathcal{C}_1\), only equivalences in \(\mathcal{C}_2\)).
For each gold clustering, gold clusters correspond to the connected components computed by only considering the (undirected) alignments of the selected alignment relations between nodes in $S$.
Hence, all alignment relations are regarded as symmetric (undirected links) and transitive (connected components), which is coherent with the majority of alignment relations (see Table~\ref{tab:clustering-links}).
Figure~\ref{fig:gold-clusters} presents the sizes of the resulting gold clusters.
We notice that many gold clusters have a size lower or equal to 10, and that considering \texttt{skos:related} or \texttt{skos:broadMatch} links increases the maximal size of gold clusters.
The availability of all these different alignment relations also allows to perform the distance analysis described in Subsection~\ref{subsection:distance-analysis} and indicated in Figure~\ref{fig:approach-outline}.

\begin{table*}
	\caption{Alignment relations considered in each gold clustering to compute the gold clusters used in our experiments. 
		We indicate whether a relation is transitive (T or $\neg$ T) and symmetric (S or $\neg$ S).}
	\label{tab:clustering-links}
	\begin{center}
		\begin{small}
			\begin{tabular}{cccccc}
				\hline
				& \texttt{owl:sameAs} & \texttt{skos:closeMatch} & \texttt{skos:relatedMatch} & \texttt{skos:related} & \texttt{skos:broadMatch} \\
				& T, S & T, S & T, S & $\neg$ T, S & T, $\neg$ S \\
				\hline
				$\mathcal{C}_0$ & $\times$ & $\times$ & $\times$ & $\times$ & $\times$ \\
				$\mathcal{C}_1$ & $\times$ & $\times$ & $\times$ & $\times$ & \\
				$\mathcal{C}_2$ & $\times$ & & & & \\
				$\mathcal{C}_3$ & & $\times$ & & & \\
				$\mathcal{C}_4$ & & & $\times$ & & \\
				$\mathcal{C}_5$ & & & & $\times$ & \\
				$\mathcal{C}_6$ & & & & & $\times$\\
				\hline
			\end{tabular}
		\end{small}
	\end{center}
\end{table*}

\subsection{Learning node embeddings}
\label{subsection:node-embeddings}

We experimented our approach with different pairs $(\mathcal{C}_i, \mathcal{G}_j)$ that were selected for their experimental interest.
All gold clusterings were experimented with graphs $\mathcal{G}_0$ and $\mathcal{G}_5$ to have a global view of the impact on performance of applying inference rules associated with domain knowledge.
All graphs were experimented with $\mathcal{C}_0$ to have a finer evaluation of each inference rule on the most heterogeneous gold clustering.
For each pair $(\mathcal{C}_i, \mathcal{G}_j)$, a 5-fold cross-validation was performed as follows.
For each $\mathcal{C}_i$, $S$ is split into five sets $S^i_k$ ($k \in \left\{1,2,3,4,5\right\}$).
All $S^i_k$ contain the same number of nodes for each gold cluster of $\mathcal{C}_i$ larger than 5 nodes.
Each set $S^i_k$ is successively used as the test set $S_{\text{test}}$, while set $S^i_{(k + 1)}$ is used as the validation set $S_{\text{val}}$\footnote{$S^i_1$ is the validation set when $S_{\text{test}} = S^i_5$.}.
Remaining sets form the train set $S_{\text{train}}$.
This corresponds to a random split of nodes to match with 60\% train -- 20\% validation -- 20\% test.

An architecture formed by 3 GCN layers is used to learn node embeddings.
The input layer consists of a featureless approach as in~\cite{kipfW17,schlichtkrullKB18}, \textit{i.e.}, the input is just a one-hot vector for each node of the graph.
It should be noted that this one-hot vector encoding can scale to relatively large knowledge graphs because \textit{(i)} we use look-up mechanisms in weight matrices based on node indices, and thus one-hot vectors are actually never stored in memory or used in computations, and \textit{(ii)} we use a basis-decomposition to limit the number of parameters.
All three layers of our architecture have an output dimension of 16.
Therefore, output embeddings for all nodes in the knowledge graph are in $\mathbb{R}^{16}$. 
The activation function used on the input and hidden layers is $\mathrm{tanh}$ while the output layer uses a linear function.
We use a basis-decomposition of 10 bases and set $c_{i,r} = |\mathcal{N}_i^r|$ for all $i$ and all $r$.
Our architecture is similar to the one used in~\cite{schlichtkrullKB18} except for the dimension of the hidden layer and the activation function of the output layer.
In such a 3-layer architecture, it follows from Eq.~\eqref{eq:gcn} that only neighboring nodes up to 3 hops\footnote{The 3-hop neighborhood of a node \(n\) consists of all the nodes that can be reached with a breadth-first traversal that starts at \(n\) and traverses at most 3 edges.} of nodes in $S$ will have an impact on their embeddings, output at layer 3.
Thus, to save memory, we reduce graphs to such 3-hop neighborhoods.	
Statistics about these reduced graphs are available in Table~\ref{tab:graph-statistics}.

Only the embeddings of nodes in $S$ (here, the PGx relationships) are considered in our clustering task.
Hence, only these embeddings are constrained in the SNN loss.
However, in $\lsnn$ (Eq.~\eqref{eq:snn-loss}), each node needs at least one other node assigned to the same gold cluster (\textit{i.e.}, having the same label).
Thus, only gold clusters of size greater or equal to 10 are used in the learning process since each $S^i_k$ contains at least 2 nodes of these clusters.
This is particularly needed for the validation and test losses but we chose to use the same constraint for the train loss for homogeneity.
We use the Adam optimizer~\cite{kingmaB14} with a starting learning rate of $0.01$.
$T$ is initialized to 1.
We learn during 200 epochs with an early-stopping mechanism: if the validation loss does not decrease of $0.0001$ after $10$ epochs, the learning process is stopped.

\begin{table}
	\caption{Statistics of PGxLOD and its transformations as described in Section~\ref{section:reasoning-mechanisms}.
		Statistics for PGxLOD discard literals and edges incident to literals.
		As we use a 3-layer architecture, statistics for all $\mathcal{G}_i$ only consider neighboring nodes up to 3 hops of nodes in $S$ (\textit{i.e.}, PGx relationships to match).
		\# denotes ``number of''.}
	\label{tab:graph-statistics}
	\begin{center}
		\begin{tabular}{cccc}
			\hline
			& \# nodes & \# edges & \# predicates \\
			\hline
			PGxLOD & 11,808,396 & 43,341,712 & 416 \\
			$\mathcal{G}_0$ & 3,758,814 & 39,956,844 & 689 \\
			$\mathcal{G}_1$ & 3,879,081 & 46,960,365 & 733 \\
			$\mathcal{G}_2$ & 3,758,814 & 22,085,701 & 347 \\
			$\mathcal{G}_3$ & 3,758,814 & 41,048,190 & 697 \\
			$\mathcal{G}_4$ & 3,758,928 & 42,691,984 & 701 \\
			$\mathcal{G}_5$ & 3,882,945 & 27,277,789 & 375 \\
			\hline
		\end{tabular}
	\end{center}
\end{table}

\subsection{Clustering}

Clustering algorithms are only applied on the embeddings of nodes in $S_{\text{test}}$ since they are the nodes we aim to match.
Recall that the learning process only considers nodes belonging to gold clusters whose size is greater or equal to 10.
Accordingly, we apply the three clustering algorithms introduced in Table~\ref{tab:clustering-algorithms} and evaluate their performance on embeddings of nodes in $S_{\text{test}}$ that belong to gold clusters whose size is greater or equal to 50, 20, and 10.
These different sizes allow to evaluate the influence of inference rules in the performance of our matching approach when considering only large or all gold clusters.

Results on all gold clusterings and graphs $\mathcal{G}_0$ and $\mathcal{G}_5$ are summarized in Table~\ref{tab:results-all-clusterings}.
Detailed results are available Table~\ref{tab:results-g0-g5-50}, Table~\ref{tab:results-g0-g5-20} and Table~\ref{tab:results-g0-g5-10} in the Appendix.
In these supplementary tables, gray cells indicate the best results among clustering algorithms given a gold clustering, a graph, and a metric.
For example, in Table~\ref{tab:results-g0-g5-50}, considering \(\mathcal{C}_0\) and $\mathcal{G}_0$, the best ACC is obtained with the Single clustering algorithm.
Underlined values indicate the best result between $\mathcal{G}_0$ and $\mathcal{G}_5$ given a gold clustering and a metric.
For example, in Table~\ref{tab:results-g0-g5-50}, given \(\mathcal{C}_1\), the best NMI for Ward is obtained with \(\mathcal{G}_0\) whereas the best ACC is obtained with \(\mathcal{G}_5\).
We notice in Table~\ref{tab:results-all-clusterings} that applying all inference rules (\textit{i.e.}, $\mathcal{G}_5$) generally increases performance for $\mathcal{C}_0$ and $\mathcal{C}_1$ which are gold clusterings that mix different alignment relations.
Results for the other gold clusterings do not show such an homogeneous and important increase in performance.

\begin{table}
	\caption{Summary of the results of clustering nodes for all gold clusterings and graphs \(\mathcal{G}_0\) and \(\mathcal{G}_5\).
		For each gold clustering, a cross (\(\times\)) indicates whether the best results were obtained with \(\mathcal{G}_0\) or \(\mathcal{G}_5\).
		Detailed results are available in Tables~\ref{tab:results-g0-g5-50}, \ref{tab:results-g0-g5-20}, and~\ref{tab:results-g0-g5-10} in the
		Appendix.}
	\label{tab:results-all-clusterings}
	\centering
	\begin{tabular}{cccc}
		
		Size of gold clusters & Gold clustering & $\mathcal{G}_0$ & $\mathcal{G}_5$\\
		\hline
		\multirow{7}*{$\geq 50$} & $\mathcal{C}_0$ & & \(\times\) \\
		& $\mathcal{C}_1$ & & \(\times\) \\
		& $\mathcal{C}_2$ & \(\times\) & \\
		& $\mathcal{C}_3$ & \(\times\) & \(\times\) \\
		& $\mathcal{C}_4$ & \(\times\) & \(\times\) \\
		& $\mathcal{C}_5$ & \(\times\) & \\
		& $\mathcal{C}_6$ & \(\times\) & \\
		\hline
		\multirow{7}*{$\geq 20$} & $\mathcal{C}_0$ & & \(\times\) \\
		& $\mathcal{C}_1$ & & \(\times\) \\
		& $\mathcal{C}_2$ & & \(\times\) \\
		& $\mathcal{C}_3$ & \(\times\) & \\
		& $\mathcal{C}_4$ & & \(\times\) \\
		& $\mathcal{C}_5$ & \(\times\) & \\
		& $\mathcal{C}_6$ & \(\times\) & \(\times\) \\
		\hline
		\multirow{7}*{$\geq 10$} & $\mathcal{C}_0$ & & \(\times\) \\
		& $\mathcal{C}_1$ & & \(\times\) \\
		& $\mathcal{C}_2$ & \(\times\) & \(\times\) \\
		& $\mathcal{C}_3$ & \(\times\) & \\
		& $\mathcal{C}_4$ & & \(\times\) \\
		& $\mathcal{C}_5$ & \(\times\) & \(\times\) \\
		& $\mathcal{C}_6$ & & \(\times\) \\
		\hline
	\end{tabular}
\end{table}

\begin{table}
	\caption{Summary of the results of our clustering for \(\mathcal{C}_0\) and all graphs.
		Detailed results are available
		in Tables~\ref{tab:results-m0-50}, \ref{tab:results-m0-20}, and~\ref{tab:results-m0-10} in the
		Appendix.}
	\label{tab:results-all-graphs}
	\begin{center}
		\begin{tabular}{ll}
			\hline
			Graph &  Performance \\
			\hline
			$\mathcal{G}_0$ & Baseline \\
			$\mathcal{G}_1$ & Improvements \\
			$\mathcal{G}_2$ & Light deterioration\\
			$\mathcal{G}_3$ & Improvements \\
			$\mathcal{G}_4$ & Consistent deterioration \\
			$\mathcal{G}_5$ & Improvements -- Best results\\
			\hline
		\end{tabular}
	\end{center}
\end{table}

Results on $\mathcal{C}_0$ and all graphs are summarized in Table~\ref{tab:results-all-graphs}.
Detailed results are available in Table~\ref{tab:results-m0-50}, Table~\ref{tab:results-m0-20}, and Table~\ref{tab:results-m0-10} in the Appendix.
In these tables, gray cells indicate the best result among clustering algorithms and underlined values indicate the best result between graphs.
For example, in Table~\ref{tab:results-m0-50}, given \(\mathcal{G}_0\), the best ACC is obtained with the Single clustering algorithm.
Given the Single algorithm, the best ARI is obtained with \(\mathcal{G}_3\) and \(\mathcal{G}_5\).
Here again, we notice that applying all inference rules (\textit{i.e.}, \(\mathcal{G}_5\)) leads to the best results.
However, computing all instantiations based on the transitive closure of the subsumption (\textit{i.e.}, \(\mathcal{G}_4\)) seems to degrade clustering performance.

\subsection{Distance analysis}
\label{subsection:distance-analysis}

\begin{figure*}
	\begin{center}
		\begin{subfigure}{0.19\textwidth}
			\begin{center}
				\includegraphics[width=\textwidth]{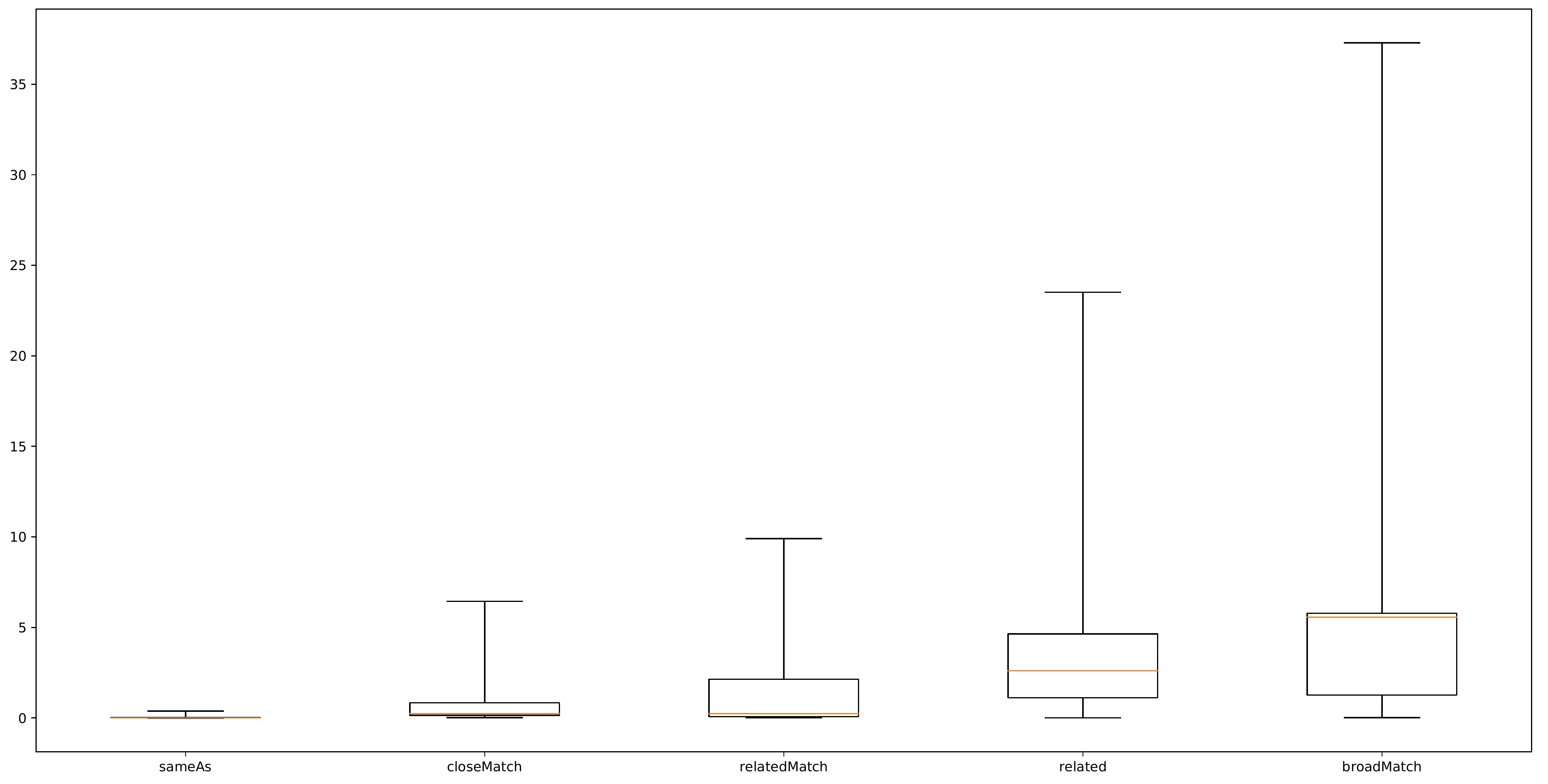}
			\end{center}
			\caption{$\mathcal{G}_0$ -- Fold 1}
		\end{subfigure}
		\begin{subfigure}{0.19\textwidth}
			\begin{center}
				\includegraphics[width=\textwidth]{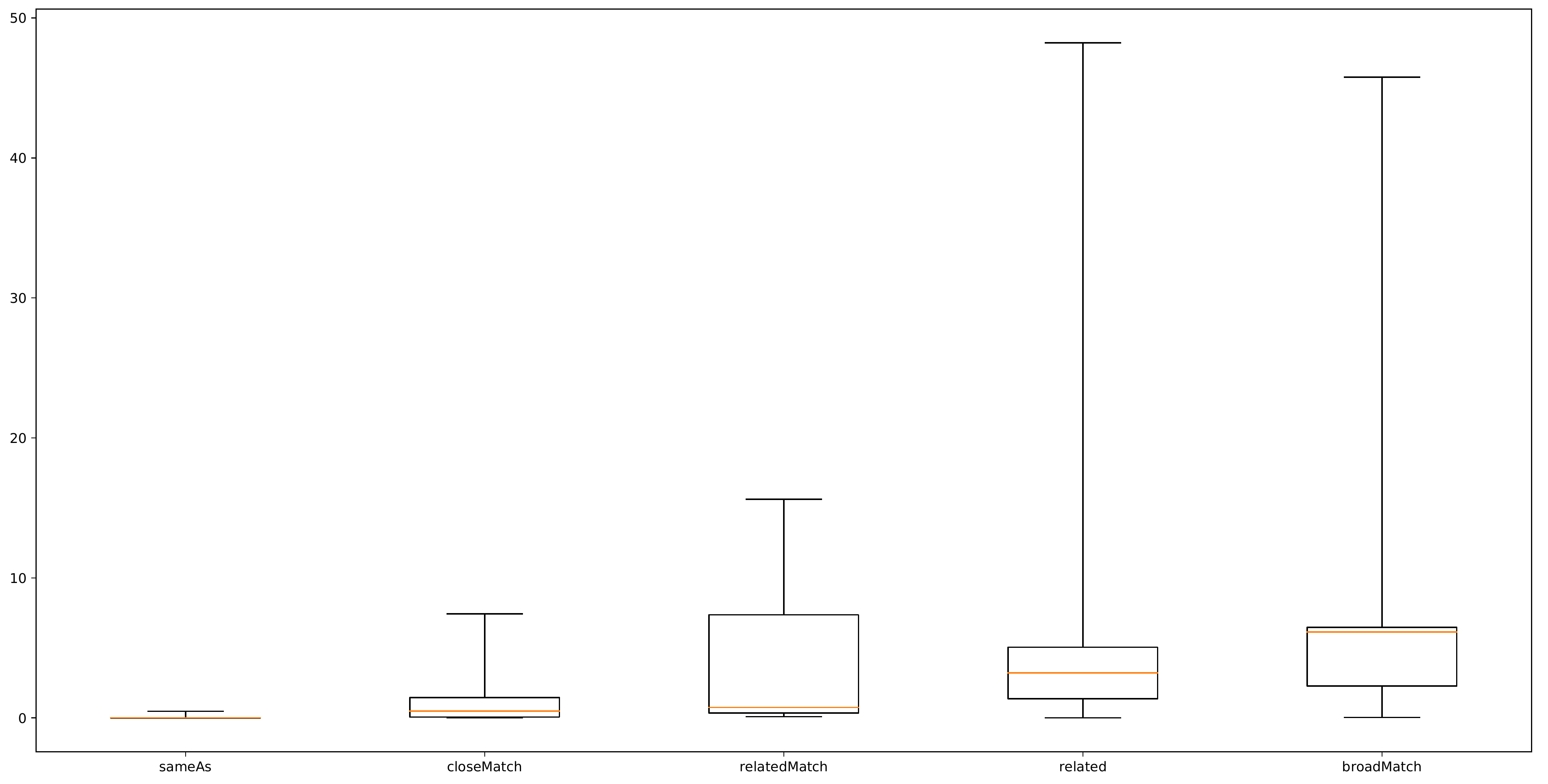}
			\end{center}
			\caption{$\mathcal{G}_0$ -- Fold 2}
		\end{subfigure}
		\begin{subfigure}{0.19\textwidth}
			\begin{center}
				\includegraphics[width=\textwidth]{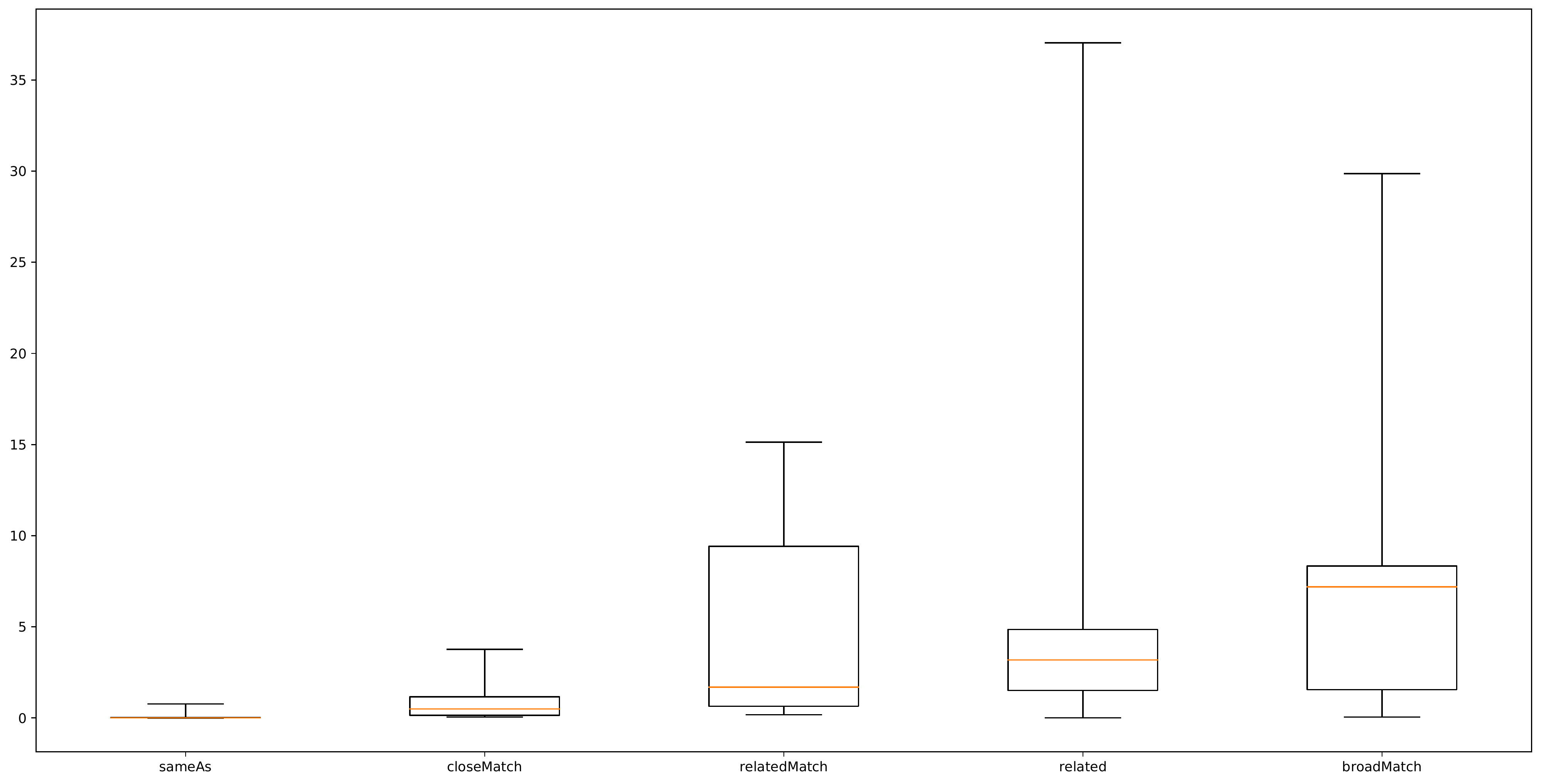}
			\end{center}
			\caption{$\mathcal{G}_0$ -- Fold 3}
		\end{subfigure}
		\begin{subfigure}{0.19\textwidth}
			\begin{center}
				\includegraphics[width=\textwidth]{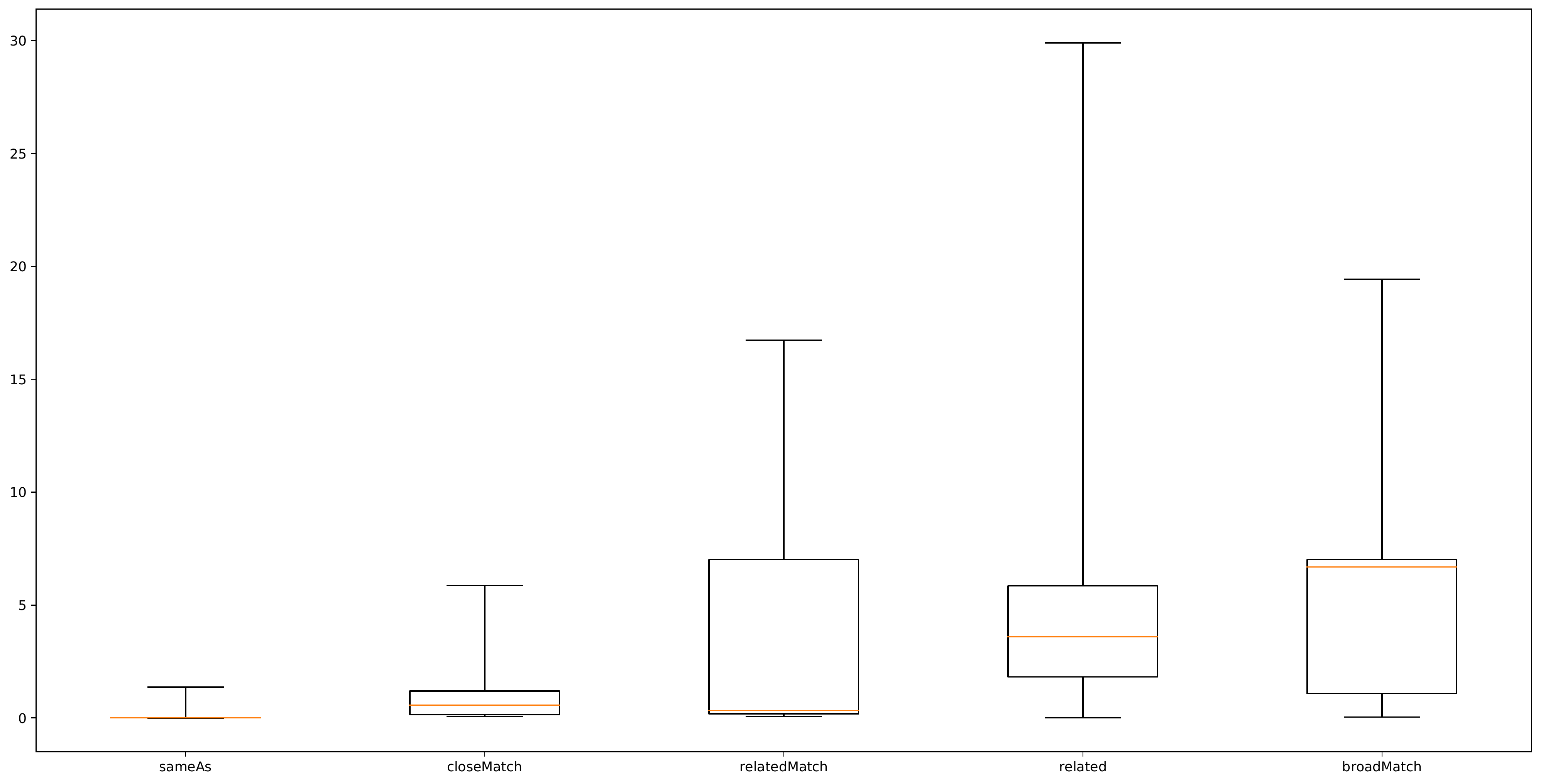}
			\end{center}
			\caption{$\mathcal{G}_0$ -- Fold 4}
		\end{subfigure}
		\begin{subfigure}{0.19\textwidth}
			\begin{center}
				\includegraphics[width=\textwidth]{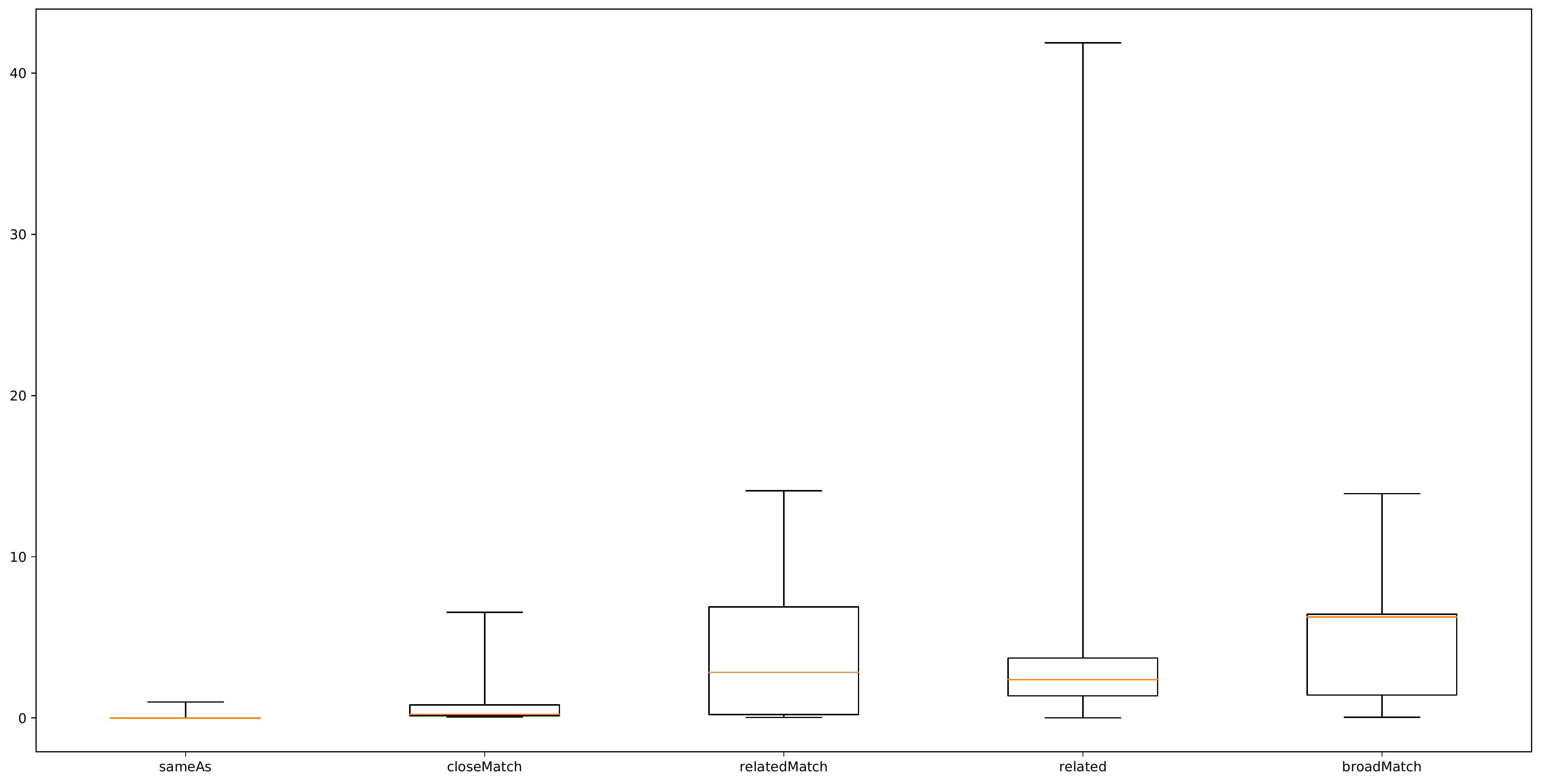}
			\end{center}
			\caption{$\mathcal{G}_0$ -- Fold 5}
		\end{subfigure}	
		
		\vspace{2em}
		
		\begin{subfigure}{0.19\textwidth}
			\begin{center}
				\includegraphics[width=\textwidth]{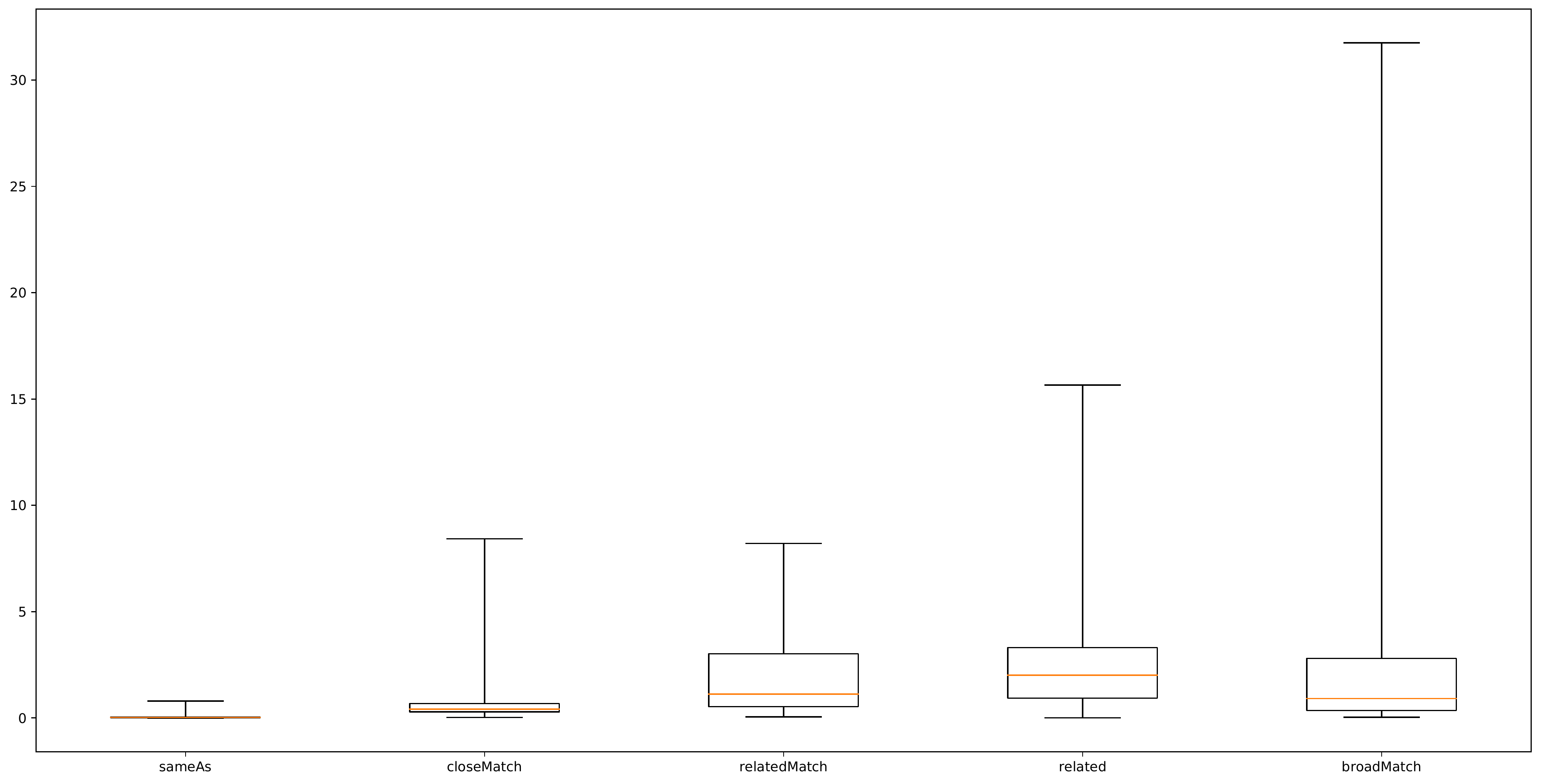}
			\end{center}
			\caption{$\mathcal{G}_5$ -- Fold 1}
		\end{subfigure}
		\begin{subfigure}{0.19\textwidth}
			\begin{center}
				\includegraphics[width=\textwidth]{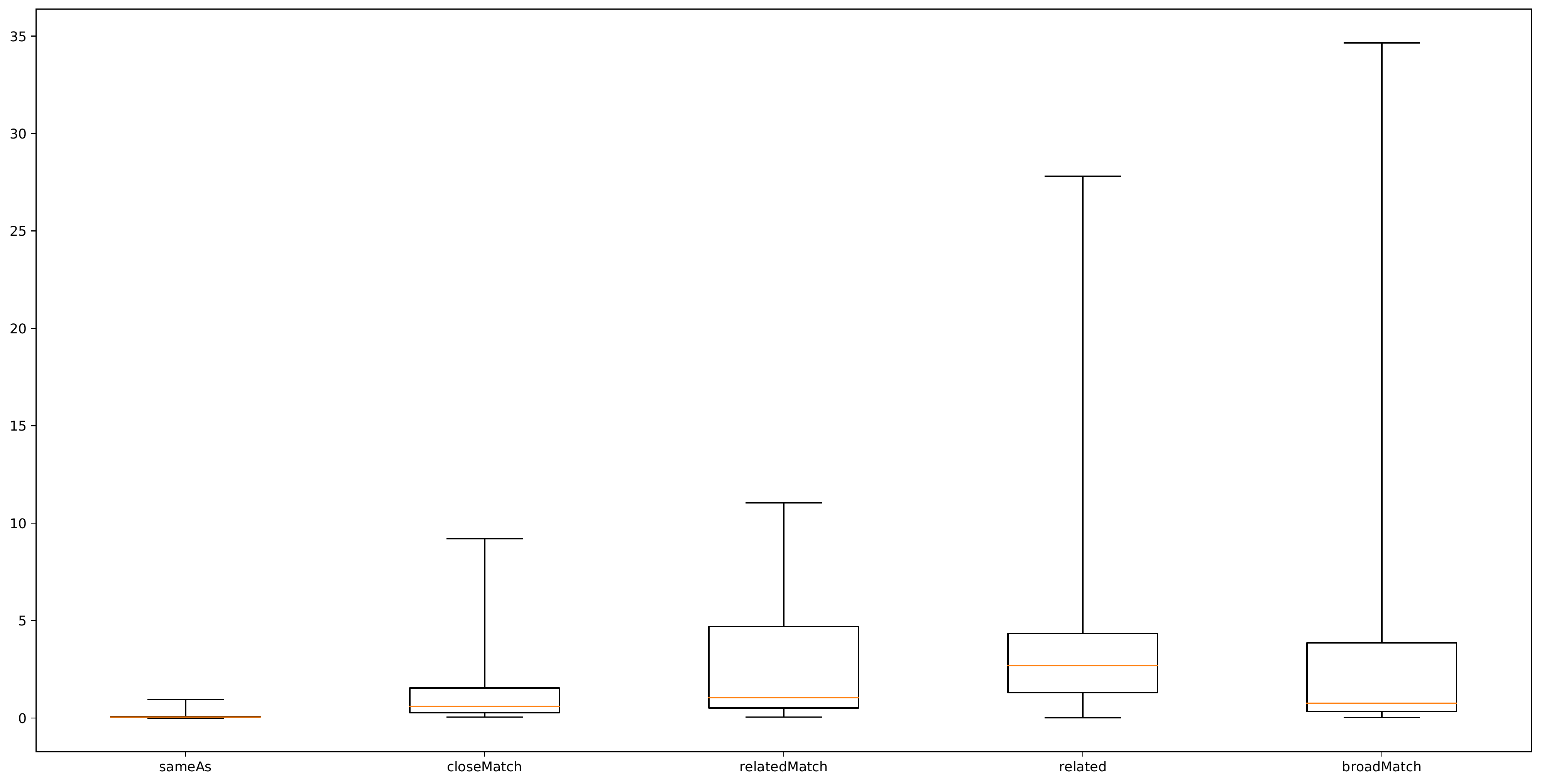}
			\end{center}
			\caption{$\mathcal{G}_5$ -- Fold 2}
		\end{subfigure}
		\begin{subfigure}{0.19\textwidth}
			\begin{center}
				\includegraphics[width=\textwidth]{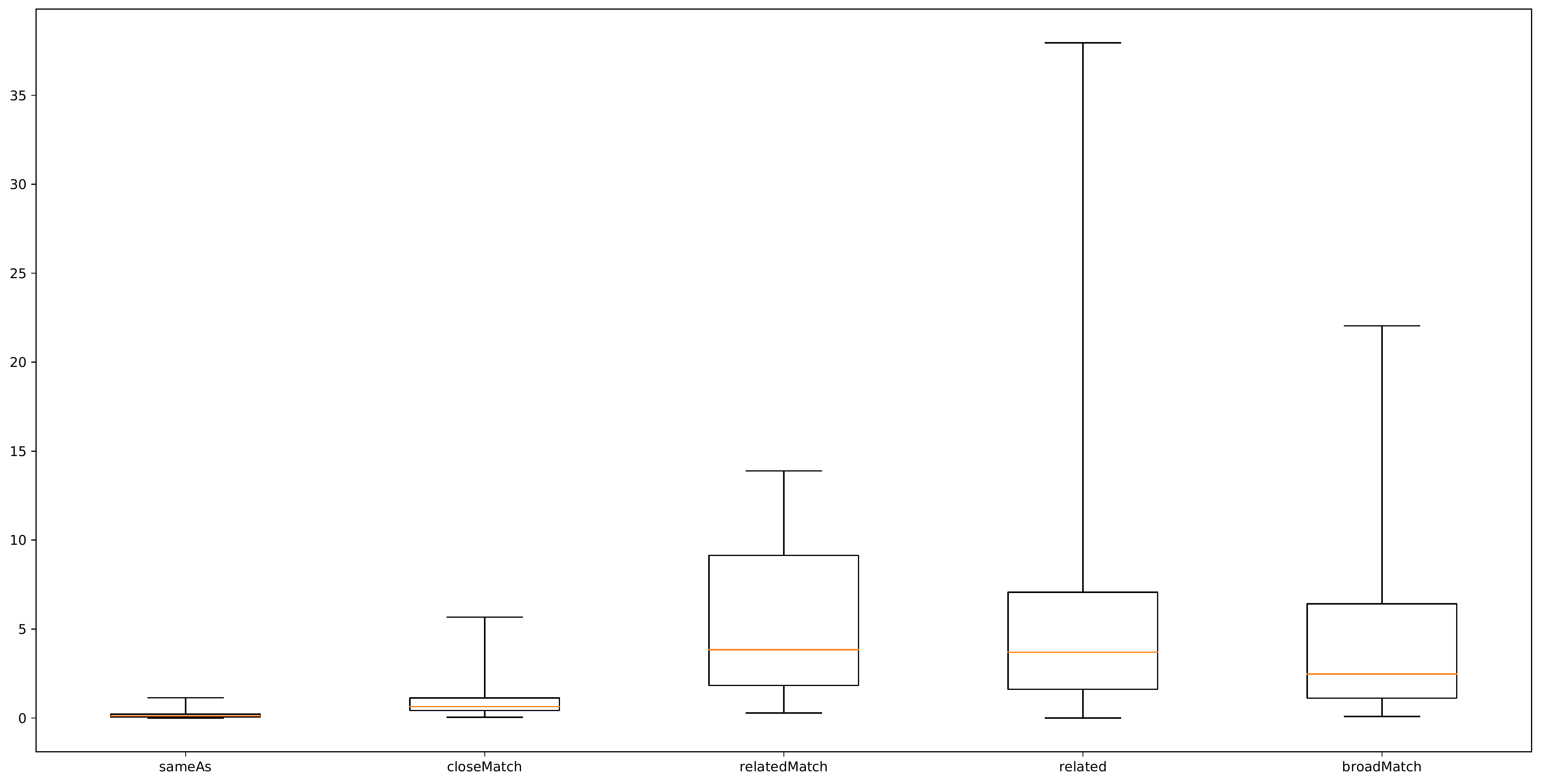}
			\end{center}
			\caption{$\mathcal{G}_5$ -- Fold 3}
		\end{subfigure}
		\begin{subfigure}{0.19\textwidth}
			\begin{center}
				\includegraphics[width=\textwidth]{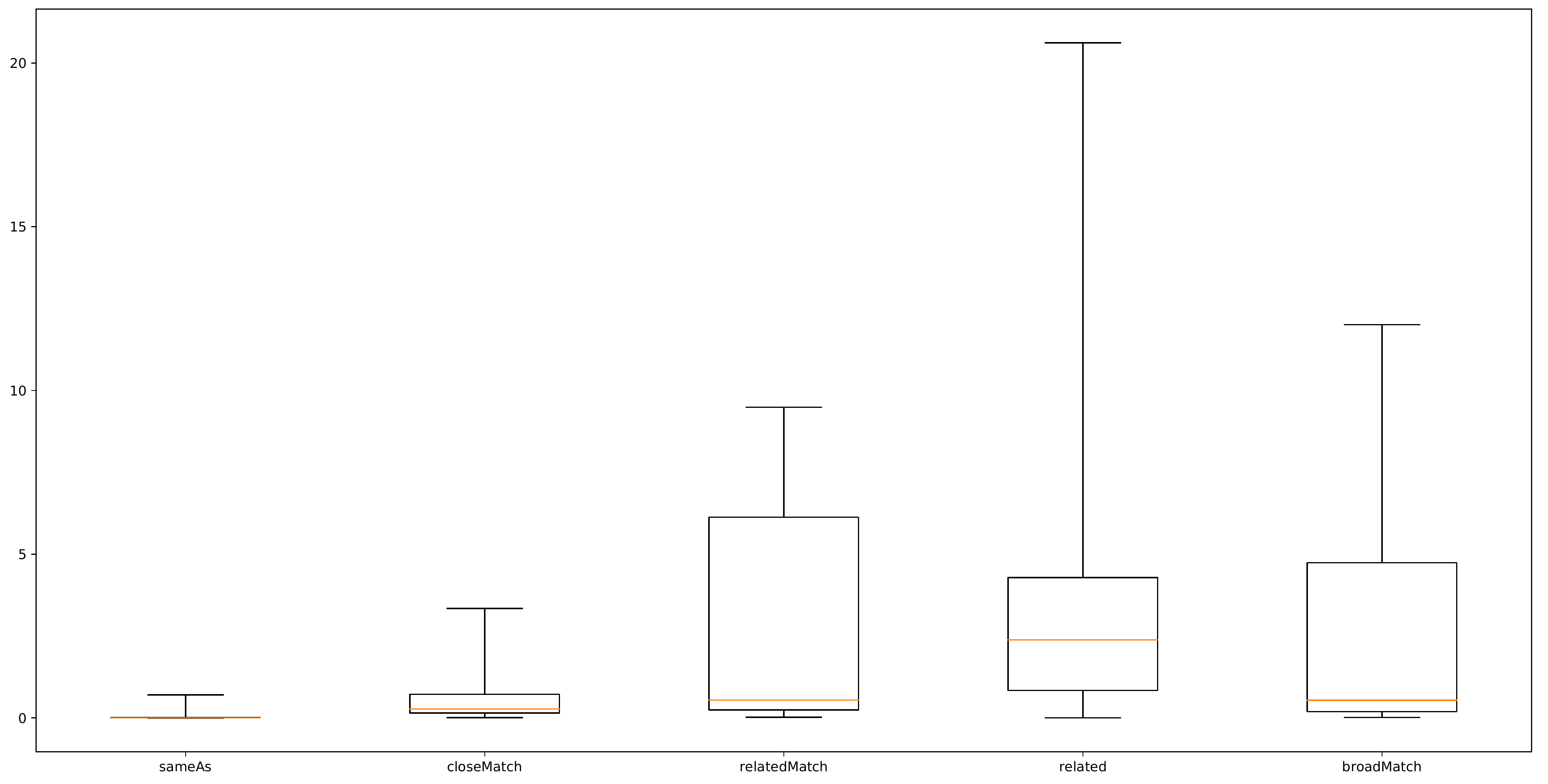}
			\end{center}
			\caption{$\mathcal{G}_5$ -- Fold 4}
		\end{subfigure}
		\begin{subfigure}{0.19\textwidth}
			\begin{center}
				\includegraphics[width=\textwidth]{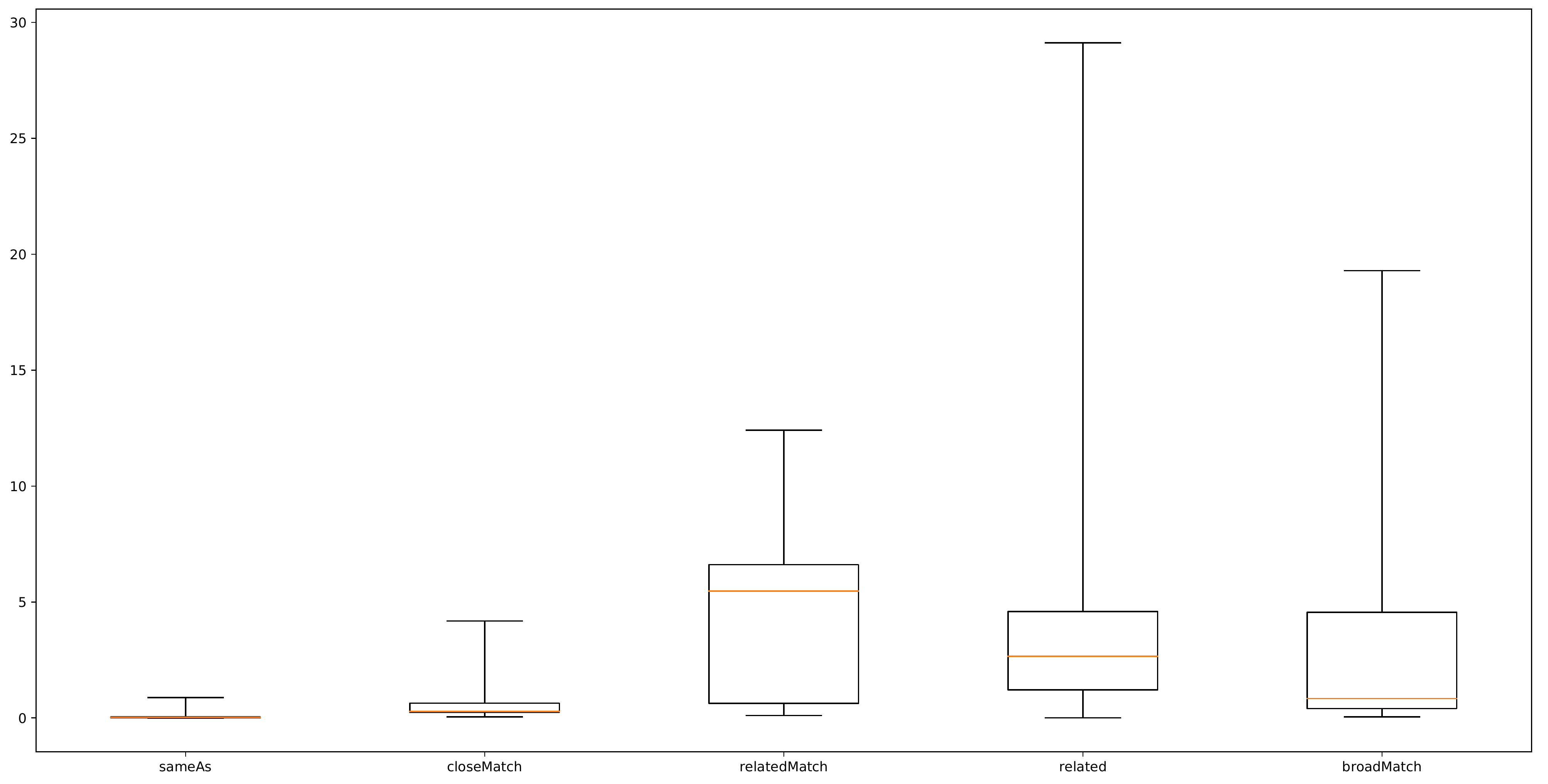}
			\end{center}
			\caption{$\mathcal{G}_5$ -- Fold 5}
		\end{subfigure}	
	\end{center}
	\caption{Distributions of distances between similar nodes by alignment relation for each test set, the $\mathcal{C}_0$ gold clustering and the two graphs $\mathcal{G}_0$ and $\mathcal{G}_5$.
		In each subpicture, links are from left to right: \texttt{owl:sameAs}, \texttt{skos:closeMatch}, \texttt{skos:relatedMatch}, \texttt{skos:related}, and \texttt{skos:broadMatch}.
	}
	\label{fig:distance-analysis-m0}
\end{figure*}

During learning and clustering, our model is unaware of the different alignment relations holding between similar nodes.
Indeed, the SNN loss only considers labels of gold clusters that do not indicate the alignment relations used to compute these clusters.
This is particularly relevant for gold clusterings $\mathcal{C}_0$ and $\mathcal{C}_1$ that mix different alignment relations to compute the gold clusters.
However, inspired by our preliminary results~\cite{monninRNC19}, we display in Figure~\ref{fig:distance-analysis-m0} the distributions of distances between similar nodes in the test set by alignment relation.
This analysis is presented for $\mathcal{C}_0$ and graphs $\mathcal{G}_0$ and $\mathcal{G}_5$.
Interestingly, similarly to our preliminary results~\cite{monninRNC19}, such distributions of distances are coherent with the ``strength'' of the alignment relations.
Indeed, for example, nodes that are weakly similar tend to be further apart than equivalent nodes.
Only the \texttt{skos:broadMatch} relation presents different distance distributions with regard to the distance distributions of the other relations across the different test sets.
This could be explained since this is the only non-symmetric relation (see Table~\ref{tab:clustering-links}).

\section{Discussion}
\label{section:discussion}

\subsection{Impact of inference rules}

It appears that $\mathcal{G}_5$ (\textit{i.e.}, all inference rules) consistently increases  performance for $\mathcal{C}_0$ and $\mathcal{C}_1$  (see Table~\ref{tab:results-all-clusterings}).
Other gold clusterings ($\mathcal{C}_2-\mathcal{C}_6$) do not show such an homogeneous and important increase in performance between $\mathcal{G}_0$ and $\mathcal{G}_5$.
$\mathcal{C}_0$ and $\mathcal{C}_1$ mix different alignment relations, which leads to a more difficult matching task.
This let us think that inference rules associated with domain knowledge provide useful improvements when dealing with heterogeneous similarities and clusters.

In $\mathcal{C}_0$ (clustering with all alignment relations)
inference rules seem to improve results with most of the expended graphes, except for $\mathcal{G}_4$ (see Table~\ref{tab:results-all-graphs}).
In  $\mathcal{G}_4$, class instantiation is satured.
Consequently, ``general'' classes are directly linked to entities that instantiate them instead of indirectly.
Hence, when computing the embeddings of such entities, embeddings of both general and specific classes are directly considered through the same predicate \texttt{type}, which makes difficult for the GCN to weight these classes differently.
As specific classes are more important than general classes to discriminate similar and dissimilar nodes, their undifferentiated influence in $\mathcal{G}_4$ embeddings may explain the decrease in performance.
We notice that $\mathcal{G}_5$ performs best, which advocates for considering all inference rules together.
However, based on the degraded performance of $\mathcal{G}_4$ with regard to \(\mathcal{G}_0\), one may want to test the progressive addition of inference from $\mathcal{G}_1$, $\mathcal{G}_2$, and $\mathcal{G}_3$.
We leave such additional experiment for future works.

\subsection{Impact of clustering set-up}

Clustering performances are generally better for gold clusterings $\mathcal{C}_2$ to $\mathcal{C}_6$ than $\mathcal{C}_0$ and $\mathcal{C}_1$ (see 
Table~\ref{tab:results-g0-g5-50}, Table~\ref{tab:results-g0-g5-20}, and Table~\ref{tab:results-g0-g5-10} in the Appendix).
These two gold clusterings mix different alignment relations when computing gold clusters, and thus their matching task is expected to be more difficult.
We also notice that performances tend to decrease when considering additional gold clusters (\textit{i.e.}, when decreasing their minimum size).
Here again, such a task is more difficult.
Indeed, clustering algorithms need to find more clusters (for Ward and Single), or clusters with a reduced minimum size (for OPTICS).
However, this is not the case for $\mathcal{C}_2$, $\mathcal{C}_3$, and $\mathcal{C}_4$.
This can be explained because, for such gold clusterings, only few gold clusters have a size greater or equal to 50 or 20 (see Figure~\ref{fig:gold-clusters}), and thus only few training examples are available.
Hence, reducing the minimum size leads to consider more training examples, and, despite the task being more difficult, improves performance.

Among the considered clustering algorithms, Single generally performs better than the others.
For $\mathcal{C}_0$ and $\mathcal{C}_1$, OPTICS is the second best algorithm.
For the other gold clusterings, Single and Ward give the best performance.
In particular, we notice that OPTICS tends to have a decent ACC but reduced ARI and NMI.
As this algorithm is unaware of the number of clusters to find and only knows their minimum size, low ARI and NMI may indicate a different clustering output in terms of both number and size of clusters.
Indeed, ARI counts the pairs of nodes that have similar of different assignments both in predicted and gold clusters and NMI measures the mutual information between two different clusterings.
On the contrary, ACC counts the number of nodes correctly assigned.
Hence, big gold clusters (partially) correctly assigned may increase the ACC value even between different clusterings.
Such a situation arises here since some of our gold clusterings lead to gold clusters with numerous nodes.
For example, Figure~\ref{fig:gold-clusters} shows that a gold cluster in $\mathcal{C}_0$ contains 17,568 nodes. 
We noticed that in some cases Ward have poor performances. The Ward algorithm minimizes the sum of squared differences within all clusters, whereas Single minimizes the distance between the closest observations of merged clusters. This let us think that the merging criterion of Single is better adapted to the loss function we use (\textit{i.e.}, the Soft Nearest Neighbor loss).

\subsection{Distance analysis}

Regarding the distance analysis of node embeddings, Figure~\ref{fig:distance-analysis-m0} shows that distances between similar nodes are different depending on the alignment relation holding between them.
Recall that our GCN model is agnostic to these alignment relations when computing the SNN loss.
Interestingly, distances reflect the ``strength'' of the alignment relations: strong similarities (\textit{i.e.}, \texttt{owl:\-same\-As} and \texttt{skos:\-close\-Match} links) have smaller distances than weaker ones (\textit{i.e.}, \texttt{skos:\-related\-Match} and \texttt{skos:related} links).
The \texttt{skos:broadMatch} relation appears more difficult to position with regard to others.
This can be explained as it is the only alignment relation that is not symmetric.
Such coherent distributions of distances seem to indicate the ``rediscovery'' of alignment relations by GCNs and encourages to consider the distance between embeddings of nodes in a ``semantic'' way, \textit{i.e.}, smaller distances indicate stronger similarities.
Hence, an interesting perspective lies in predicting the exact alignment relation holding between similar nodes (\textit{i.e.}, in the same cluster) based on the distance between their embeddings, and evaluating this prediction.
Additionally, such different distances also seem to confirm that the neighborhood aggregation of embeddings in GCNs makes them well-suited to a structural and relational matching.

\subsection{Towards a further integration of domain knowledge in GCNs}

Our results highlight the interest of considering domain knowledge associated with knowledge graphs in embedding approaches and seem to advocate for a further integration of domain knowledge within embedding models.
Future works may investigate the same targets with additional inferences rules (\textit{e.g.}, from OWL 2 RL semantics) or different embedding techniques, whether based on graph neural networks~\cite{dwivedi2020} or others (\textit{e.g.}, translational approaches such as TransE).
Additionally, we did not use attention mechanisms, which could also consider domain knowledge as in Logic Attention Network~\cite{wangHLP19}.
Here, inference rules associated with domain knowledge are used to transform the knowledge graph as a pre-processing operation.
However, we could envision to consider such mechanisms directly in the model (\textit{e.g.}, weight sharing between predicates and their super-predicates).
Literals could also be taken into account~\cite{wangLLZ18}.
In a larger perspective, one major future work lies in investigating if and how other semantics than types of alignments can emerge in the output embedding space.

\subsection{Generalization to other knowledge graphs}
Despite our approach being motivated by the matching of individuals within an aggregated knowledge graph, its transposition to distinct graphs could be explored. 
Such a perspective could allow to assess the generalization of our approach and its results.
In particular, we could consider knowledge graphs that are not completely independent such as LOD datasets that are connected through major LOD hubs. 
Recall that our approach is supervised since gold clusters are computed from preexisting alignments.
Hence, testing our approach on different knowledge graphs of the LOD would require such preexisting alignments or using ontology alignment systems in a distant supervision process~\cite{chenJHAHL21}.
In this setting, merging the different graphs into one and learning a ``global'' embedding, as we did, may provide positive results but may pose additional scalability issues.

\section{Conclusion}
\label{section:conclusion}

In this paper, we proposed to match entities of a knowledge graph by learning node embeddings with Graph Convolutional Networks (GCNs) and clustering nodes based on their embeddings.
We particularly investigated the interplay between formal semantics associated with knowledge graphs and GCN models.
Our results showed that considering inference rules associated with domain knowledge tends to improve performance.
Additionally, even if our GCN model was agnostic to the exact alignment relations holding between entities (\textit{e.g.}, equivalence, weak similarity), distances in the embedding space are coherent with the ``strength'' of the alignment relations.
These results seem to advocate for a further integration of formal semantics within embedding models.

\section*{Acknowledgments}
	This work was supported by the \textit{PractiKPharma} project, founded by the French National Research Agency (ANR) under Grant ANR15-CE23-0028, and by the \textit{Snowball} Inria Associate Team.

\bibliographystyle{unsrt}
\bibliography{bibliography}

\appendix

\section{Detailed results of clustering experiments}
\label{appendix:detailed-results}

Detailed results of clustering experiments are available Table~\ref{tab:results-g0-g5-50}, Table~\ref{tab:results-g0-g5-20} and Table~\ref{tab:results-g0-g5-10}.

\begin{table*}
	\caption{Results of clustering nodes that belong to gold clusters whose size is greater or equal to 50 for graphs $\mathcal{G}_0$ and $\mathcal{G}_5$. 
		Average and standard deviation for each metric are computed on test folds during a 5-fold cross validation.
		Given a gold clustering, gray cells indicate the best results among clustering algorithms and underlined values indicate the best result between $\mathcal{G}_0$ and $\mathcal{G}_5$.
	}
	\label{tab:results-g0-g5-50}
	\begin{center}
		\begin{tiny}
			\begin{tabular}{cc|ccc|ccc}
				\hline
				& & \multicolumn{3}{c|}{$\mathcal{G}_0$} &  \multicolumn{3}{c}{$\mathcal{G}_5$} \\
				& & ACC & ARI & NMI & ACC & ARI & NMI \\
				\hline
				\multirow{3}*{$\mathcal{C}_0$}& Ward & $0.24 \pm 0.02$ & $0.07 \pm 0.01$ & $\underline{0.37 \pm 0.01}$ & $\underline{0.25 \pm 0.02}$ & $0.07 \pm 0.01$ & $0.35 \pm 0.01$ \\
				& Single & \cellcolor{lgray}$0.84 \pm 0.08$ & \cellcolor{lgray}$0.66 \pm 0.13$ & \cellcolor{lgray}$0.59 \pm 0.05$ & \cellcolor{lgray}$\underline{0.90 \pm 0.00}$ & \cellcolor{lgray}$\underline{0.75 \pm 0.01}$ & \cellcolor{lgray}$\underline{0.64 \pm 0.02}$ \\
				& OPTICS & $0.61 \pm 0.05$ & $0.21 \pm 0.08$ & $0.25 \pm 0.04$ & $\underline{0.68 \pm 0.02}$ & $\underline{0.27 \pm 0.05}$ & $\underline{0.27 \pm 0.02}$ \\
				\hline
				\multirow{3}*{$\mathcal{C}_1$}& Ward & $0.19 \pm 0.02$ & $0.05 \pm 0.00$ & $\underline{0.33 \pm 0.01}$ & $\underline{0.20 \pm 0.03}$ & $0.05 \pm 0.01$ & $0.31 \pm 0.01$ \\
				& Single & \cellcolor{lgray}$0.85 \pm 0.01$ & \cellcolor{lgray}$0.55 \pm 0.04$ & \cellcolor{lgray}$0.51 \pm 0.03$ & \cellcolor{lgray}$0.85 \pm 0.01$ & \cellcolor{lgray}$\underline{0.57 \pm 0.03}$ & \cellcolor{lgray}$0.51 \pm 0.03$ \\
				& OPTICS & $0.64 \pm 0.03$ & $0.19 \pm 0.03$ & $0.28 \pm 0.01$ & $\underline{0.71 \pm 0.04}$ & $\underline{0.26 \pm 0.06}$ & $\underline{0.30 \pm 0.02}$ \\
				\hline
				\multirow{3}*{$\mathcal{C}_2$}& Ward & $0.88 \pm 0.03$ & $0.84 \pm 0.03$ & $0.94 \pm 0.01$ & $0.88 \pm 0.03$ & $0.84 \pm 0.03$ & $0.94 \pm 0.01$ \\
				& Single & $\underline{0.88 \pm 0.03}$ & $\underline{0.84 \pm 0.03}$ & $\underline{0.94 \pm 0.01}$ & $0.86 \pm 0.04$ & $0.81 \pm 0.07$ & $0.93 \pm 0.02$ \\
				& OPTICS & \cellcolor{lgray}$\underline{0.94 \pm 0.06}$ & \cellcolor{lgray}$\underline{0.92 \pm 0.07}$ & \cellcolor{lgray}$\underline{0.97 \pm 0.03}$ & \cellcolor{lgray}$0.91 \pm 0.06$ & \cellcolor{lgray}$0.88 \pm 0.08$ & \cellcolor{lgray}$0.95 \pm 0.04$ \\
				\hline
				\multirow{3}*{$\mathcal{C}_3$}& Ward & $0.52 \pm 0.01$ & $0.00 \pm 0.00$ & $0.00 \pm 0.00$ & $\underline{0.53 \pm 0.01}$ & $0.00 \pm 0.00$ & $0.00 \pm 0.00$ \\
				& Single & $0.53 \pm 0.01$ & $0.00 \pm 0.00$ & $0.00 \pm 0.00$ & $0.53 \pm 0.01$ & $0.00 \pm 0.00$ & $0.00 \pm 0.00$ \\
				& OPTICS & \cellcolor{lgray}$1.00 \pm 0.00$ & \cellcolor{lgray}$1.00 \pm 0.00$ & \cellcolor{lgray}$1.00 \pm 0.00$ & \cellcolor{lgray}$1.00 \pm 0.00$ & \cellcolor{lgray}$1.00 \pm 0.00$ & \cellcolor{lgray}$1.00 \pm 0.00$ \\
				\hline
				\multirow{3}*{$\mathcal{C}_4$}& Ward & \cellcolor{lgray}$0.94 \pm 0.00$ & \cellcolor{lgray}$0.00 \pm 0.00$ & $0.00 \pm 0.00$ & \cellcolor{lgray}$0.94 \pm 0.00$ & $0.00 \pm 0.00$ & $0.00 \pm 0.00$ \\
				& Single & \cellcolor{lgray}$0.94 \pm 0.00$ & \cellcolor{lgray}$0.00 \pm 0.00$ & $0.00 \pm 0.00$ & \cellcolor{lgray}$0.94 \pm 0.00$ & $0.00 \pm 0.00$ & $0.00 \pm 0.00$ \\
				& OPTICS & $\underline{0.45 \pm 0.14}$ & $-0.02 \pm 0.06$ & \cellcolor{lgray}$0.08 \pm 0.08$ & $0.38 \pm 0.11$ & \cellcolor{lgray}$\underline{0.01 \pm 0.05}$ & \cellcolor{lgray}$\underline{0.11 \pm 0.08}$ \\
				\hline
				\multirow{3}*{$\mathcal{C}_5$}& Ward & $\underline{0.20 \pm 0.02}$ & $\underline{0.04 \pm 0.00}$ & $\underline{0.24 \pm 0.01}$ & $0.15 \pm 0.02$ & $0.03 \pm 0.00$ & $0.20 \pm 0.01$ \\
				& Single & \cellcolor{lgray}$0.88 \pm 0.00$ & \cellcolor{lgray}$\underline{0.31 \pm 0.03}$ & \cellcolor{lgray}$\underline{0.29 \pm 0.03}$ & \cellcolor{lgray}$\underline{0.89 \pm 0.00}$ & \cellcolor{lgray}$0.30 \pm 0.03$ & \cellcolor{lgray}$0.27 \pm 0.02$ \\
				& OPTICS & $\underline{0.71 \pm 0.03}$ & $\underline{0.18 \pm 0.07}$ & $\underline{0.18 \pm 0.04}$ & $0.68 \pm 0.06$ & $0.07 \pm 0.07$ & $0.11 \pm 0.04$ \\
				\hline
				\multirow{3}*{$\mathcal{C}_6$}& Ward & $0.69 \pm 0.02$ & $0.48 \pm 0.03$ & \cellcolor{lgray}$0.64 \pm 0.03$ & $\underline{0.76 \pm 0.12}$ & \cellcolor{lgray}$\underline{0.58 \pm 0.22}$ & \cellcolor{lgray}$\underline{0.67 \pm 0.12}$ \\
				& Single & \cellcolor{lgray}$\underline{0.86 \pm 0.02}$ & \cellcolor{lgray}$\underline{0.60 \pm 0.09}$ & $\underline{0.63 \pm 0.08}$ & \cellcolor{lgray}$0.82 \pm 0.02$ & $0.46 \pm 0.12$ & $0.52 \pm 0.11$ \\
				& OPTICS & $\underline{0.59 \pm 0.07}$ & $\underline{0.21 \pm 0.09}$ & $0.44 \pm 0.06$ & $0.58 \pm 0.05$ & $0.19 \pm 0.06$ & $\underline{0.45 \pm 0.04}$ \\
				\hline
			\end{tabular}
		\end{tiny}
	\end{center}
\end{table*}

\begin{table}
	\caption{Results of clustering nodes that belong to gold clusters whose size is greater or equal to 20 for graphs $\mathcal{G}_0$ and $\mathcal{G}_5$. 
		Average and standard deviation for each metric are computed on test folds during a 5-fold cross validation.
		Given a gold clustering, gray cells indicate the best results among clustering algorithms and underlined values indicate the best result between $\mathcal{G}_0$ and $\mathcal{G}_5$.
	}
	\label{tab:results-g0-g5-20}
	\begin{center}
		\begin{tiny}
			\begin{tabular}{cc|ccc|ccc}
				\hline
				& & \multicolumn{3}{c|}{$\mathcal{G}_0$} &  \multicolumn{3}{c}{$\mathcal{G}_5$} \\
				& & ACC & ARI & NMI & ACC & ARI & NMI \\
				\hline
				\multirow{3}*{$\mathcal{C}_0$}& Ward & $0.17 \pm 0.01$ & $0.04 \pm 0.00$ & $\underline{0.32 \pm 0.01}$ & $0.17 \pm 0.02$ & $0.04 \pm 0.00$ & $0.31 \pm 0.01$ \\
				& Single & \cellcolor{lgray}$0.79 \pm 0.08$ & \cellcolor{lgray}$0.64 \pm 0.11$ & \cellcolor{lgray}$0.54 \pm 0.05$ & \cellcolor{lgray}$\underline{0.86 \pm 0.01}$ & \cellcolor{lgray}$\underline{0.69 \pm 0.01}$ & \cellcolor{lgray}$\underline{0.57 \pm 0.01}$ \\
				& OPTICS & $0.45 \pm 0.03$ & $0.09 \pm 0.02$ & $0.17 \pm 0.01$ & $\underline{0.50 \pm 0.01}$ & $\underline{0.13 \pm 0.01}$ & $\underline{0.19 \pm 0.01}$ \\
				\hline
				\multirow{3}*{$\mathcal{C}_1$}& Ward & $0.15 \pm 0.01$ & $0.03 \pm 0.00$ & $\underline{0.31 \pm 0.01}$ & $0.15 \pm 0.01$ & $0.03 \pm 0.00$ & $0.30 \pm 0.00$ \\
				& Single & \cellcolor{lgray}$0.64 \pm 0.22$ & \cellcolor{lgray}$0.38 \pm 0.19$ & \cellcolor{lgray}$0.45 \pm 0.06$ & \cellcolor{lgray}$\underline{0.82 \pm 0.01}$ & \cellcolor{lgray}$\underline{0.58 \pm 0.03}$ & \cellcolor{lgray}$\underline{0.52 \pm 0.03}$ \\
				& OPTICS & $0.47 \pm 0.02$ & $0.08 \pm 0.01$ & $0.20 \pm 0.01$ & $\underline{0.51 \pm 0.02}$ & $\underline{0.11 \pm 0.03}$ & $0.20 \pm 0.01$ \\
				\hline
				\multirow{3}*{$\mathcal{C}_2$}& Ward & \cellcolor{lgray}$0.98 \pm 0.00$ & \cellcolor{lgray}$0.98 \pm 0.02$ & \cellcolor{lgray}$0.99 \pm 0.00$ & \cellcolor{lgray}$0.98 \pm 0.00$ & \cellcolor{lgray}$\underline{0.99 \pm 0.01}$ & \cellcolor{lgray}$0.99 \pm 0.00$ \\
				& Single & $0.97 \pm 0.03$ & $0.95 \pm 0.05$ & $0.98 \pm 0.01$ & \cellcolor{lgray}$\underline{0.98 \pm 0.00}$ & $\underline{0.98 \pm 0.01}$ & \cellcolor{lgray}$\underline{0.99 \pm 0.00}$ \\
				& OPTICS & $0.69 \pm 0.01$ & $0.44 \pm 0.04$ & $0.78 \pm 0.01$ & $\underline{0.73 \pm 0.03}$ & $\underline{0.48 \pm 0.04}$ & $\underline{0.81 \pm 0.02}$ \\
				\hline
				\multirow{3}*{$\mathcal{C}_3$}& Ward & \cellcolor{lgray}$\underline{0.92 \pm 0.06}$ & \cellcolor{lgray}$\underline{0.89 \pm 0.08}$ & \cellcolor{lgray}$\underline{0.95 \pm 0.03}$ & $0.89 \pm 0.05$ & $0.84 \pm 0.08$ & $0.93 \pm 0.03$  \\
				& Single & $\underline{0.91 \pm 0.05}$ & $\underline{0.87 \pm 0.06}$ & \cellcolor{lgray}$\underline{0.95 \pm 0.03}$ & $0.88 \pm 0.07$ & $0.84 \pm 0.09$ & $0.93 \pm 0.04$ \\
				& OPTICS & $0.89 \pm 0.07$ & $0.87 \pm 0.08$ & $0.94 \pm 0.08$ & \cellcolor{lgray}$\underline{0.92 \pm 0.06}$ & \cellcolor{lgray}$\underline{0.90 \pm 0.09}$ & \cellcolor{lgray}$\underline{0.95 \pm 0.04}$ \\
				\hline
				\multirow{3}*{$\mathcal{C}_4$}& Ward & \cellcolor{lgray}$0.94 \pm 0.00$ & $0.00 \pm 0.00$ & $0.00 \pm 0.00$ & \cellcolor{lgray}$0.94 \pm 0.00$ & $0.00 \pm 0.00$ & $0.00 \pm 0.00$ \\
				& Single & \cellcolor{lgray}$0.94 \pm 0.00$ & $0.00 \pm 0.00$ & $0.00 \pm 0.00$ & \cellcolor{lgray}$0.94 \pm 0.00$ & $0.00 \pm 0.00$ & $0.00 \pm 0.00$ \\
				& OPTICS & $0.29 \pm 0.05$ & \cellcolor{lgray}$0.01 \pm 0.01$ & \cellcolor{lgray}$0.09 \pm 0.02$ & $\underline{0.34 \pm 0.05}$ & \cellcolor{lgray}$\underline{0.03 \pm 0.01}$ & \cellcolor{lgray}$\underline{0.11 \pm 0.02}$ \\
				\hline
				\multirow{3}*{$\mathcal{C}_5$}& Ward & $\underline{0.12 \pm 0.01}$ & $\underline{0.02 \pm 0.00}$ & $\underline{0.21 \pm 0.01}$ & $0.10 \pm 0.00$ & $0.01 \pm 0.00$ & $0.17 \pm 0.00$ \\
				& Single & \cellcolor{lgray}$0.85 \pm 0.01$ & \cellcolor{lgray}$\underline{0.32 \pm 0.09}$ & \cellcolor{lgray}$\underline{0.28 \pm 0.06}$ & \cellcolor{lgray}$\underline{0.86 \pm 0.00}$ & \cellcolor{lgray}$0.20 \pm 0.03$ & \cellcolor{lgray}$0.27 \pm 0.02$ \\
				& OPTICS & $0.48 \pm 0.02$ & $0.05 \pm 0.01$ & $\underline{0.10 \pm 0.01}$ & $\underline{0.52 \pm 0.02}$ & $0.05 \pm 0.02$ & $0.09 \pm 0.01$ \\
				\hline
				\multirow{3}*{$\mathcal{C}_6$}& Ward & $\underline{0.56 \pm 0.05}$ & $\underline{0.39 \pm 0.10}$ & \cellcolor{lgray}$\underline{0.67 \pm 0.02}$ & $0.50 \pm 0.06$ & $0.29 \pm 0.08$ & $0.65 \pm 0.03$ \\
				& Single & \cellcolor{lgray}$0.64 \pm 0.07$ & \cellcolor{lgray}$0.43 \pm 0.13$ & $0.62 \pm 0.05$ & \cellcolor{lgray}$\underline{0.78 \pm 0.01}$ & \cellcolor{lgray}$\underline{0.67 \pm 0.06}$ & \cellcolor{lgray}$\underline{0.71 \pm 0.03}$ \\
				& OPTICS & $0.44 \pm 0.03$ & $0.08 \pm 0.03$ & $\underline{0.38 \pm 0.02}$ & $\underline{0.47 \pm 0.05}$ & $0.08 \pm 0.08$ & $0.37 \pm 0.05$ \\
				\hline
			\end{tabular}
		\end{tiny}
	\end{center}
\end{table}

\begin{table}
	\caption{Results of clustering nodes that belong to gold clusters whose size is greater or equal to 10 for graphs $\mathcal{G}_0$ and $\mathcal{G}_5$. 
		Average and standard deviation for each metric are computed on test folds during a 5-fold cross validation.
		Given a gold clustering, gray cells indicate the best results among clustering algorithms and underlined values indicate the best result between $\mathcal{G}_0$ and $\mathcal{G}_5$.
	}
	\label{tab:results-g0-g5-10}
	\begin{center}
		\begin{tiny}
			\begin{tabular}{cc|ccc|ccc}
				\hline
				& & \multicolumn{3}{c|}{$\mathcal{G}_0$} &  \multicolumn{3}{c}{$\mathcal{G}_5$} \\
				& & ACC & ARI & NMI & ACC & ARI & NMI \\
				\hline
				\multirow{3}*{$\mathcal{C}_0$}& Ward & $\underline{0.14 \pm 0.01}$ & $0.02 \pm 0.00$ & $\underline{0.29 \pm 0.01}$ & $0.13 \pm 0.01$ & $0.02 \pm 0.00$ & $0.28 \pm 0.02$ \\
				& Single & \cellcolor{lgray}$0.66 \pm 0.17$ & \cellcolor{lgray}$0.53 \pm 0.22$ & \cellcolor{lgray}$0.52 \pm 0.06$ & \cellcolor{lgray}$\underline{0.74 \pm 0.15}$ & \cellcolor{lgray}$\underline{0.61 \pm 0.16}$ & \cellcolor{lgray}$\underline{0.54 \pm 0.06}$ \\
				& OPTICS & $0.25 \pm 0.02$ & $0.02 \pm 0.01$ & $\underline{0.12 \pm 0.01}$ & $\underline{0.27 \pm 0.01}$ & $\underline{0.03 \pm 0.01}$ & $0.11 \pm 0.01$ \\
				\hline
				\multirow{3}*{$\mathcal{C}_1$}& Ward & $0.13 \pm 0.01$ & $0.01 \pm 0.00$ & $\underline{0.28 \pm 0.01}$ & $\underline{0.14 \pm 0.01}$ & $0.01 \pm 0.00$ & $0.27 \pm 0.01$ \\
				& Single & \cellcolor{lgray}$0.41 \pm 0.12$ & \cellcolor{lgray}$0.18 \pm 0.07$ & \cellcolor{lgray}$0.41 \pm 0.02$ & \cellcolor{lgray}$\underline{0.72 \pm 0.15}$ & \cellcolor{lgray}$\underline{0.53 \pm 0.14}$ & \cellcolor{lgray}$\underline{0.52 \pm 0.04}$ \\
				& OPTICS & $0.28 \pm 0.01$ & $0.02 \pm 0.00$ & $0.13 \pm 0.01$ & $0.28 \pm 0.00$ & $0.02 \pm 0.00$ & $0.13 \pm 0.01$ \\
				\hline
				\multirow{3}*{$\mathcal{C}_2$}& Ward & \cellcolor{lgray}$0.99 \pm 0.00$ & \cellcolor{lgray}$0.99 \pm 0.00$ & \cellcolor{lgray}$0.99 \pm 0.00$ & \cellcolor{lgray}$0.99 \pm 0.00$ & \cellcolor{lgray}$0.99 \pm 0.00$ & \cellcolor{lgray}$0.99 \pm 0.00$ \\
				& Single & \cellcolor{lgray}$0.99 \pm 0.01$ & \cellcolor{lgray}$0.99 \pm 0.02$ & \cellcolor{lgray}$0.99 \pm 0.00$ & \cellcolor{lgray}$0.99 \pm 0.00$ & \cellcolor{lgray}$0.99 \pm 0.00$ & \cellcolor{lgray}$0.99 \pm 0.00$ \\
				& OPTICS & $\underline{0.63 \pm 0.01}$ & $\underline{0.31 \pm 0.01}$ & $\underline{0.67 \pm 0.01}$ & $0.62 \pm 0.01$ & $0.29 \pm 0.01$ & $0.66 \pm 0.01$ \\
				\hline
				\multirow{3}*{$\mathcal{C}_3$}& Ward & \cellcolor{lgray}$\underline{0.92 \pm 0.00}$ & \cellcolor{lgray}$\underline{0.90 \pm 0.10}$ & \cellcolor{lgray}$\underline{0.94 \pm 0.05}$ & \cellcolor{lgray}$0.86 \pm 0.04$ & \cellcolor{lgray}$0.81 \pm 0.05$ & \cellcolor{lgray}$0.89 \pm 0.02$ \\
				& Single & $\underline{0.90 \pm 0.07}$ & $\underline{0.88 \pm 0.12}$ & $\underline{0.93 \pm 0.05}$ & $0.83 \pm 0.05$ & $0.77 \pm 0.08$ & $0.88 \pm 0.04$ \\
				& OPTICS & $\underline{0.75 \pm 0.02}$ & $\underline{0.58 \pm 0.03}$ & $\underline{0.78 \pm 0.02}$ & $0.72 \pm 0.03$ & $0.49 \pm 0.07$ & $0.73 \pm 0.05$ \\
				\hline
				\multirow{3}*{$\mathcal{C}_4$}& Ward & \cellcolor{lgray}$0.99 \pm 0.00$ & \cellcolor{lgray}$0.90 \pm 0.07$ & \cellcolor{lgray}$0.86 \pm 0.08$ & \cellcolor{lgray}$0.99 \pm 0.00$ & \cellcolor{lgray}$\underline{0.91 \pm 0.05}$ & \cellcolor{lgray}$\underline{0.88 \pm 0.04}$ \\
				& Single & $0.98 \pm 0.01$ & $0.83 \pm 0.10$ & $0.78 \pm 0.15$ & \cellcolor{lgray}$\underline{0.99 \pm 0.00}$ & $\underline{0.88 \pm 0.05}$ & $\underline{0.85 \pm 0.07}$ \\
				& OPTICS & $0.18 \pm 0.03$ & $\underline{0.01 \pm 0.01}$ & $0.06 \pm 0.01$ & $0.18 \pm 0.01$ & $0.00 \pm 0.00$ & $0.06 \pm 0.01$ \\
				\hline
				\multirow{3}*{$\mathcal{C}_5$}& Ward & $\underline{0.09 \pm 0.01}$ & $0.01 \pm 0.00$ & $\underline{0.18 \pm 0.01}$ & $0.07 \pm 0.00$ & $0.01 \pm 0.00$ & $0.14 \pm 0.01$ \\
				& Single & \cellcolor{lgray}$0.81 \pm 0.01$ & \cellcolor{lgray}$0.31 \pm 0.12$ & \cellcolor{lgray}$0.25 \pm 0.08$ & \cellcolor{lgray}$\underline{0.82 \pm 0.01}$ & \cellcolor{lgray}$\underline{0.32 \pm 0.08}$ & \cellcolor{lgray}$\underline{0.26 \pm 0.05}$ \\
				& OPTICS & $0.27 \pm 0.01$ & $0.01 \pm 0.00$ & $\underline{0.06 \pm 0.01}$ & $\underline{0.28 \pm 0.01}$ & $0.01 \pm 0.01$ & $0.05 \pm 0.01$ \\
				\hline
				\multirow{3}*{$\mathcal{C}_6$}& Ward & $\underline{0.48 \pm 0.03}$ & $\underline{0.24 \pm 0.05}$ & $\underline{0.64 \pm 0.01}$ & $0.44 \pm 0.02$ & $0.16 \pm 0.03$ & $0.60 \pm 0.02$ \\
				& Single & \cellcolor{lgray}$0.63 \pm 0.07$ & \cellcolor{lgray}$0.56 \pm 0.14$ & \cellcolor{lgray}$0.70 \pm 0.04$ & \cellcolor{lgray}$\underline{0.74 \pm 0.02}$ & \cellcolor{lgray}$\underline{0.76 \pm 0.05}$ & \cellcolor{lgray}$\underline{0.76 \pm 0.03}$ \\
				& OPTICS & $0.37 \pm 0.02$ & $0.02 \pm 0.01$ & $0.29 \pm 0.02$ & $0.37 \pm 0.02$ & $\underline{0.03 \pm 0.01}$ & $0.29 \pm 0.01$ \\
				\hline
			\end{tabular}
		\end{tiny}
	\end{center}
\end{table}

\begin{table}
	\caption{Results of clustering nodes that belong to gold clusters whose size is greater or equal to 50 for $\mathcal{C}_0$ and all graphs.
		Average and standard deviation for each metric are computed on test folds during a 5-fold cross validation.
		Gray cells indicate the best result among clustering algorithms.
		Underlined values indicate the best result between graphs.
		$\downarrow$ indicates a lower value with regard to $\mathcal{G}_0$.
	}
	\label{tab:results-m0-50}
	\begin{center}
		\begin{tiny}
			\begin{tabular}{cccccccc}
				\hline
				& & \multicolumn{2}{c}{Ward} & \multicolumn{2}{c}{Single} & \multicolumn{2}{c}{OPTICS}  \\
				\hline
				\multirow{3}*{$\mathcal{G}_0$}& ACC & & $0.24 \pm 0.02$ & & \cellcolor{lgray}$0.84 \pm 0.08$ & & $0.61 \pm 0.05$ \\
				& ARI & & $0.07 \pm 0.01$ & & \cellcolor{lgray}$0.66 \pm 0.13$ & & $0.21 \pm 0.08$ \\
				& NMI & & $\underline{0.37 \pm 0.01}$ & & \cellcolor{lgray}$0.59 \pm 0.05$ & & $0.25 \pm 0.04$ \\
				\hline
				\multirow{3}*{$\mathcal{G}_1$}& ACC & & $0.24 \pm 0.02$ & & \cellcolor{lgray}$0.86 \pm 0.00$ & $\downarrow$ & $0.58 \pm 0.03$ \\
				& ARI & & $0.07 \pm 0.01$ & & \cellcolor{lgray}$0.70 \pm 0.02$ & $\downarrow$ & $0.16 \pm 0.03$ \\
				& NMI & $\downarrow$ & $0.35 \pm 0.02$ & $\downarrow$ & \cellcolor{lgray}$0.58 \pm 0.02$ & $\downarrow$ & $0.23 \pm 0.01$ \\
				\hline
				\multirow{3}*{$\mathcal{G}_2$}& ACC & & $0.24 \pm 0.01$ & & \cellcolor{lgray}$0.89 \pm 0.02$ & & $\underline{0.70 \pm 0.01}$ \\
				& ARI & & $0.07 \pm 0.00$ & & \cellcolor{lgray}$0.72 \pm 0.03$ & & $\underline{0.32 \pm 0.03}$ \\
				& NMI & $\downarrow$ & $0.34 \pm 0.01$ & & \cellcolor{lgray}$0.61 \pm 0.04$ & &$\underline{0.28 \pm 0.01}$ \\
				\hline
				\multirow{3}*{$\mathcal{G}_3$}& ACC & $\downarrow$ & $0.22 \pm 0.03$ & & \cellcolor{lgray}$0.89 \pm 0.02$ & & $0.63 \pm 0.03$ \\
				& ARI & & $0.07 \pm 0.01$ & & \cellcolor{lgray}$\underline{0.75 \pm 0.02}$ & & $0.29 \pm 0.04$ \\
				& NMI & $\downarrow$ & $0.36 \pm 0.01$ & & \cellcolor{lgray}$0.63 \pm 0.03$ & & $\underline{0.28 \pm 0.02}$ \\
				\hline
				\multirow{3}*{$\mathcal{G}_4$}& ACC & $\downarrow$ & $0.23 \pm 0.02$ & $\downarrow$ & \cellcolor{lgray}$0.80 \pm 0.16$ & & $0.62 \pm 0.02$ \\
				& ARI & & $0.07 \pm 0.01$ & $\downarrow$ & \cellcolor{lgray}$0.63 \pm 0.21$ & $\downarrow$ & $0.20 \pm 0.03$ \\
				& NMI & $\downarrow$ & $0.36 \pm 0.01$ & $\downarrow$ & \cellcolor{lgray}$0.58 \pm 0.08$ & $\downarrow$ & $0.24 \pm 0.01$ \\
				\hline
				\multirow{3}*{$\mathcal{G}_5$}& ACC & & $\underline{0.25 \pm 0.02}$ & & \cellcolor{lgray}$\underline{0.90 \pm 0.00}$ & & $0.68 \pm 0.02$ \\
				& ARI & & $0.07 \pm 0.01$ & & \cellcolor{lgray}$\underline{0.75 \pm 0.01}$ & & $0.27 \pm 0.05$ \\
				& NMI & $\downarrow$ & $0.35 \pm 0.01$ & & \cellcolor{lgray}$\underline{0.64 \pm 0.02}$ & & $0.27 \pm 0.02$ \\
				\hline
			\end{tabular}
		\end{tiny}
	\end{center}
\end{table}

\begin{table}
	\caption{Results of clustering nodes that belong to gold clusters whose size is greater or equal to 20 for $\mathcal{C}_0$ and all graphs.
		Average and standard deviation for each metric are computed on test folds during a 5-fold cross validation.
		Gray cells indicate the best result among clustering algorithms.
		Underlined values indicate the best result between graphs.
		$\downarrow$ indicates a lower value with regard to $\mathcal{G}_0$.
	}
	\label{tab:results-m0-20}
	\begin{center}
		\begin{tiny}
			\begin{tabular}{cccccccc}
				\hline
				& & \multicolumn{2}{c}{Ward} & \multicolumn{2}{c}{Single} & \multicolumn{2}{c}{OPTICS}  \\
				\hline
				\multirow{3}*{$\mathcal{G}_0$}& ACC & & $0.17 \pm 0.01$ & & \cellcolor{lgray} $0.79 \pm 0.08$ & & $0.45 \pm 0.03$ \\
				& ARI & & $\underline{0.04 \pm 0.00}$ & & \cellcolor{lgray}$0.64 \pm 0.11$ & & $0.09 \pm 0.02$ \\
				& NMI & & $\underline{0.32 \pm 0.01}$ & & \cellcolor{lgray}$0.54 \pm 0.05$ & & $0.17 \pm 0.01$ \\
				\hline
				\multirow{3}*{$\mathcal{G}_1$}& ACC & & $\underline{0.19 \pm 0.01}$ & & \cellcolor{lgray}$0.81 \pm 0.01$ & $\downarrow$ & $0.43 \pm 0.02$ \\
				& ARI & & $\underline{0.04 \pm 0.00}$ & & \cellcolor{lgray}$0.64 \pm 0.02$ & $\downarrow$ & $0.07 \pm 0.03$ \\
				& NMI & & $\underline{0.32 \pm 0.01}$ & $\downarrow$ & \cellcolor{lgray}$0.52 \pm 0.01$ & $\downarrow$ & $0.16 \pm 0.02$ \\
				\hline
				\multirow{3}*{$\mathcal{G}_2$}& ACC & & $0.17 \pm 0.01$ & & \cellcolor{lgray} $0.81 \pm 0.08$ & & $0.48 \pm 0.01$ \\
				& ARI & & $\underline{0.04 \pm 0.00}$ & $\downarrow$ & \cellcolor{lgray} $0.63 \pm 0.09$ & & $0.11 \pm 0.02$ \\
				& NMI & $\downarrow$ & $0.30 \pm 0.01$ & & \cellcolor{lgray} $0.54 \pm 0.05$ & & $0.17 \pm 0.01$ \\
				\hline
				\multirow{3}*{$\mathcal{G}_3$}& ACC & $\downarrow$ & $0.15 \pm 0.01$ & & \cellcolor{lgray} $0.81 \pm 0.06$ & & $0.46 \pm 0.01$ \\
				& ARI & $\downarrow$ & $0.03 \pm 0.00$ & & \cellcolor{lgray} $0.64 \pm 0.12$ & & $0.12 \pm 0.02$ \\
				& NMI & & $\underline{0.32 \pm 0.01}$ & & \cellcolor{lgray} $0.55 \pm 0.05$ & & $0.18 \pm 0.01$ \\
				\hline
				\multirow{3}*{$\mathcal{G}_4$}& ACC & & $0.17 \pm 0.01$ & $\downarrow$ & \cellcolor{lgray} $0.69 \pm 0.22$ & $\downarrow$ & $0.43 \pm 0.02$ \\
				& ARI & $\downarrow$ & $0.03 \pm 0.00$ & $\downarrow$ & \cellcolor{lgray} $0.54 \pm 0.24$ & $\downarrow$ & $0.08 \pm 0.02$ \\
				& NMI & & $\underline{0.32 \pm 0.01}$ & $\downarrow$ & \cellcolor{lgray} $0.52 \pm 0.08$ & & $0.17 \pm 0.01$ \\
				\hline
				\multirow{3}*{$\mathcal{G}_5$}& ACC & & $0.17 \pm 0.02$ & & \cellcolor{lgray} $\underline{0.86 \pm 0.01}$ & & $\underline{0.50 \pm 0.01}$ \\
				& ARI & & $\underline{0.04 \pm 0.00}$ & & \cellcolor{lgray} $\underline{0.69 \pm 0.01}$ & & $\underline{0.13 \pm 0.01}$\\
				& NMI & $\downarrow$ & $0.31 \pm 0.01$ & & \cellcolor{lgray} $\underline{0.57 \pm 0.01}$ & & $\underline{0.19 \pm 0.01}$ \\
				\hline
			\end{tabular}
		\end{tiny}
	\end{center}
\end{table}

\begin{table}
	\caption{Results of clustering nodes that belong to gold clusters whose size is greater or equal to 10 for $\mathcal{C}_0$ and all graphs.
		Average and standard deviation for each metric are computed on test folds during a 5-fold cross validation.
		Gray cells indicate the best result among clustering algorithms.
		Underlined values indicate the best result between graphs.
		$\downarrow$ indicates a lower value with regard to $\mathcal{G}_0$.
	}
	\label{tab:results-m0-10}
	\begin{center}
		\begin{tiny}
			\begin{tabular}{cccccccc}
				\hline
				& & \multicolumn{2}{c}{Ward} & \multicolumn{2}{c}{Single} & \multicolumn{2}{c}{OPTICS}  \\
				\hline
				\multirow{3}*{$\mathcal{G}_0$}& ACC & & $0.14 \pm 0.01$ & & \cellcolor{lgray} $0.66 \pm 0.17$ & & $0.25 \pm 0.02$ \\
				& ARI & & $0.02 \pm 0.00$ & & \cellcolor{lgray}$0.53 \pm 0.22$ & & $0.02 \pm 0.01$ \\
				& NMI & & $0.29 \pm 0.01$ & & \cellcolor{lgray}$0.52 \pm 0.06$ & & $\underline{0.12 \pm 0.01}$ \\
				\hline
				\multirow{3}*{$\mathcal{G}_1$}& ACC & & $\underline{0.15 \pm 0.01}$ & & \cellcolor{lgray}$0.73 \pm 0.10$ & & $0.25 \pm 0.01$ \\
				& ARI & & $0.02 \pm 0.00$ & & \cellcolor{lgray}$0.58 \pm 0.13$ & & $0.02 \pm 0.01$ \\
				& NMI & & $\underline{0.30 \pm 0.01}$ & $\downarrow$ & \cellcolor{lgray}$0.51 \pm 0.03$ & & $\underline{0.12 \pm 0.01}$ \\
				\hline
				\multirow{3}*{$\mathcal{G}_2$}& ACC & $\downarrow$ & $0.12 \pm 0.01$ & $\downarrow$ & \cellcolor{lgray}$0.62 \pm 0.16$ & & $\underline{0.27 \pm 0.01}$ \\
				& ARI & & $0.02 \pm 0.00$ & $\downarrow$ & \cellcolor{lgray}$0.47 \pm 0.19$ & & $\underline{0.03 \pm 0.01}$ \\
				& NMI & $\downarrow$ & $0.26 \pm 0.01$ & $\downarrow$ & \cellcolor{lgray}$0.48 \pm 0.05$ & $\downarrow$ & $0.11 \pm 0.00$ \\
				\hline
				\multirow{3}*{$\mathcal{G}_3$}& ACC & $\downarrow$ & $0.12 \pm 0.00$ & & \cellcolor{lgray}$0.70 \pm 0.18$ & & $0.26 \pm 0.01$ \\
				& ARI & & $0.02 \pm 0.00$ & & \cellcolor{lgray}$0.58 \pm 0.23$ & & $\underline{0.03 \pm 0.01}$ \\
				& NMI & $\downarrow$ & $0.28 \pm 0.01$ & & \cellcolor{lgray}$0.52 \pm 0.06$ & & $\underline{0.12 \pm 0.01}$ \\
				\hline
				\multirow{3}*{$\mathcal{G}_4$}& ACC & & $0.14 \pm 0.01$ & $\downarrow$ & \cellcolor{lgray}$0.56 \pm 0.18$ & & $0.25 \pm 0.01$ \\
				& ARI & & $0.02 \pm 0.00$ & $\downarrow$ & \cellcolor{lgray}$0.42 \pm 0.20$ & & $0.02 \pm 0.00$ \\
				& NMI & & $0.29 \pm 0.01$ & $\downarrow$ & \cellcolor{lgray}$0.50 \pm 0.06$ & & $\underline{0.12 \pm 0.00}$ \\
				\hline
				\multirow{3}*{$\mathcal{G}_5$}& ACC & $\downarrow$ & $0.13 \pm 0.01$ & & \cellcolor{lgray}$\underline{0.74 \pm 0.15}$ & & $\underline{0.27 \pm 0.01}$ \\
				& ARI & & $0.02 \pm 0.00$ & & \cellcolor{lgray}$\underline{0.61 \pm 0.16}$ & & $\underline{0.03 \pm 0.01}$ \\
				& NMI & $\downarrow$ & $0.28 \pm 0.02$ & & \cellcolor{lgray}$\underline{0.54 \pm 0.06}$ & $\downarrow$ & $0.11 \pm 0.01$ \\
				\hline
			\end{tabular}
		\end{tiny}
	\end{center}
\end{table}

\end{document}